\documentclass[sigconf]{acmart}
\DeclareMathOperator*{\argmax}{argmax}
\usepackage{hhline}
\usepackage{multirow}
\usepackage{threeparttable}
\usepackage{algorithm}
\usepackage{algorithmic}
\usepackage{array}
\usepackage{adjustbox}
\usepackage{caption}

\newcolumntype{L}[1]{>{\raggedright\let\newline\\\arraybackslash\hspace{0pt}}m{#1}}
\newcolumntype{C}[1]{>{\centering\let\newline\\\arraybackslash\hspace{0pt}}m{#1}}
\newcolumntype{R}[1]{>{\raggedleft\let\newline\\\arraybackslash\hspace{0pt}}m{#1}}

\AtBeginDocument{%
  \providecommand\BibTeX{{%
    \normalfont B\kern-0.5em{\scshape i\kern-0.25em b}\kern-0.8em\TeX}}}

\copyrightyear{2023}
\acmYear{2023}
\setcopyright{acmlicensed}\acmConference[CIKM '23]{Proceedings of the 32nd ACM International Conference on Information and Knowledge Management}{October 21--25, 2023}{Birmingham, United Kingdom}
\acmBooktitle{Proceedings of the 32nd ACM International Conference on Information and Knowledge Management (CIKM '23), October 21--25, 2023, Birmingham, United Kingdom}
\acmPrice{15.00}
\acmDOI{10.1145/3583780.3614828}
\acmISBN{979-8-4007-0124-5/23/10}

\begin{document}

\title[CGRec]{Cracking the Code of Negative Transfer: \\ A Cooperative Game Theoretic Approach for Cross-Domain Sequential Recommendation} 

\author{Chung Park}
\email{cpark88kr@gmail.com}
\affiliation{%
  \institution{SK Telecom}
  \country{ }
}
\affiliation{%
  \institution{Korea Advanced Institute of Science and Technology}
  \country{ }
}

\author{Taesan Kim}
\email{ktmountain@sk.com}
\affiliation{%
  \institution{SK Telecom}
  \country{ }
  }

\author{Taekyoon Choi}
\email{tgchoi03@gmail.com}
\affiliation{%
  \institution{NAVER}
  \country{ }
  }

\author{Junui Hong}
\email{skt.juhong@sk.com}
\affiliation{%
  \institution{SK Telecom}
  \country{ }
  }
\affiliation{%
  \institution{Korea Advanced Institute of Science and Technology}
  \country{ }
}

\author{Yelim Yu}
\email{yelim.yu@sk.com}
\affiliation{%
  \institution{SK Telecom}
  \country{ }
  }

\author{Mincheol Cho}
\email{skt.mccho@sk.com}
\affiliation{%
  \institution{SK Telecom}
  \country{ }
  }

\author{Kyunam Lee}
\email{kyunam@sk.com}
\affiliation{%
  \institution{SK Telecom}
  \country{ }
  }

\author{Sungil Ryu}
\email{ryu0121@sk.com}
\affiliation{%
  \institution{SK Telecom}
  \country{ }
  }

\author{Hyungjun Yoon}
\email{hjyoon@sk.com}
\affiliation{%
  \institution{SK Telecom}
  \country{ }
  }

\author{Minsung Choi}
\email{ms.choi@sk.com}
\affiliation{%
  \institution{SK Telecom}
  \country{ }
  }

\author{Jaegul Choo}
\authornote{Corresponding Author (jchoo@kaist.ac.kr)}
\email{jchoo@kaist.ac.kr}
\affiliation{%
  \institution{Korea Advanced Institute of Science and Technology}
  \country{ }
  }


\renewcommand{\shortauthors}{Chung Park et al.}

\begin{abstract}
This paper investigates Cross-Domain Sequential Recommendation (CDSR), a promising method that uses information from multiple domains (more than three) to generate accurate and diverse recommendations, and takes into account the sequential nature of user interactions. 
The effectiveness of these systems often depends on the complex interplay among the multiple domains. 
In this dynamic landscape, the problem of negative transfer arises, where heterogeneous knowledge between dissimilar domains leads to performance degradation due to differences in user preferences across these domains.
As a remedy, we propose a new CDSR framework that addresses the problem of negative transfer by assessing the extent of negative transfer from one domain to another and adaptively assigning low weight values to the corresponding prediction losses. 
To this end, the amount of negative transfer is estimated by measuring the marginal contribution of each domain to model performance based on a cooperative game theory.
In addition, a hierarchical contrastive learning approach that incorporates information from the sequence of coarse-level categories into that of fine-level categories (e.g., item level) when implementing contrastive learning was developed to mitigate negative transfer.
Despite the potentially low relevance between domains at the fine-level, there may be higher relevance at the category level due to its generalised and broader preferences.
We show that our model is superior to prior works in terms of model performance on two real-world datasets across ten different domains. 
\end{abstract}

\ccsdesc[500]{Information systems~Recommender systems}
\ccsdesc[500]{Computing methodologies~Neural networks}

\keywords{Cross-Domain Sequential Recommendation; Negative Transfer; Cooperative Game}


\maketitle

\section{Introduction} \label{section: introduction}
In recommendation systems, user preferences typically reflect partial or incomplete engagement within a specific domain.
This tendency results in biased recommendation outcomes based on the historical interactions observed within a single domain \cite{cao2022contrastive}.
Cross-Domain Sequential Recommendation (CDSR), which aims to enhance the performance of multiple domain recommendations simultaneously by utilizing information from other domains, has been introduced to mitigate this concern.
Another key aspect of the CDSR is capturing the sequential characteristics of the user interaction, which enables a model to better understand users' dynamic interest over time \cite{li2023one,cao2022contrastive}.

    \begin{figure}[ht]
    \begin{center}
    \includegraphics[width=1 \linewidth]{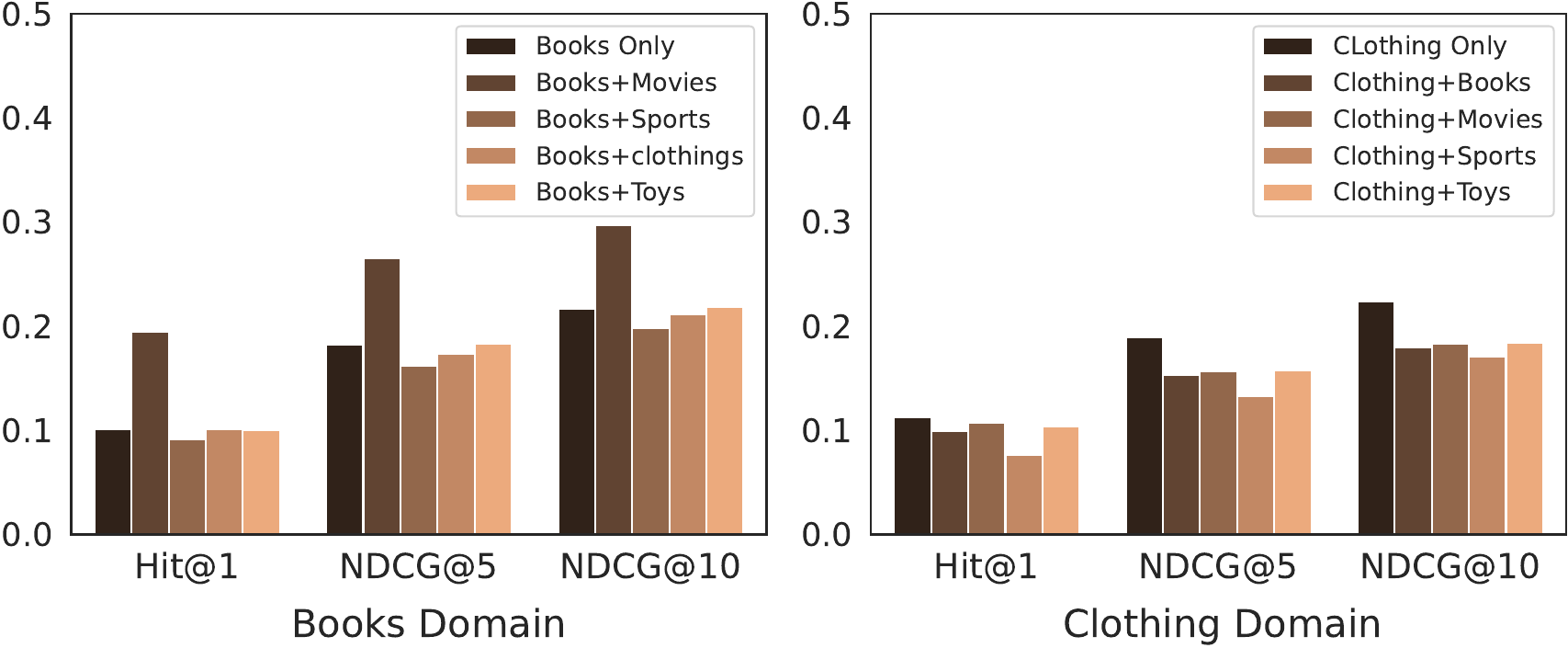}
    \end{center}
    \vspace{-0.3cm}
    \caption{
    Negative transfer cases between domain-domain pairs in the \textit{Amazon} dataset is shown.
    The figure on the left compares the single-domain recommendation results for the \textit{Books} domain with the cross-domain recommendation results for the \textit{Books} domain paired with four other domains.
    The \textit{Books}+\textit{Sports} case shows worse performance than the \textit{Books} case alone, due to the irrelevance between the \textit{Books} and \textit{Sports} domains (i.e., negative transfer).
    On the right is about \textit{Clothing} domain with the same scheme as the left side.
    The performance of both the single- and cross-domain recommendation cases was evaluated using the C$^{2}$DSR \cite{cao2022contrastive}.}
    \label{fig:collaboration_conflict_example}
    \end{figure}

In the dynamic landscape of the CDSR task, a phenomenon known as \textit{negative transfer} arises when dissimilar knowledge across domains results in a performance decline due to discrepancies in user preferences among these domains \cite{gao2013cross, li2023one, zhang2020overcoming}. 
For instance, as shown in Figure \ref{fig:collaboration_conflict_example}, the \textit{Books} domain enjoys a collaborative relationship with the \textit{Movies} domain, yet maintains conflicting relationships with the \textit{Sports} domain. This is due to the different domain interests of a same user.
Note that the \textit{Books}+\textit{Movies} case shows higher performance than the \textit{Books} only case.
From this perspective, \citet{li2023one} proposed the cross-domain recommendation method, which alleviates the negative transfer problem by training both a global and a domain-specific user representation, but their approach does not consider the sequential dynamics of user interaction.
As users' preferences change over time, the dynamics of negative transfer further intensify and this has not been fully addressed in previous studies.

In this paper, we propose the model \textbf{CGRec}, which stands for \underline{C}ooperative \underline{G}ame Theoretic Approach for Cross-Domain Sequential \underline{Rec}ommendation.
This model mitigates the negative transfer issue by minimizing its detrimental impact on the overall performance of each domain.
In particular, we focus on the case of more than three domains for real-world applications, which has not been considered much in the previous CDSR studies.
For this purpose, our model has been trained using hybrid sequences from more than three domains, conceptualizing each domain as players in a cooperative game \cite{myerson1991game}. 
Then, Shapley values \cite{straffin1993game} were derived for all domains and were used to approximate the degree of the negative transfer of each domain.
This value is used to assign low weights to items in domains that show high negative transfer during the training process.

\begin{figure}[ht]
    \begin{center}
    \includegraphics[width=0.9 \linewidth]{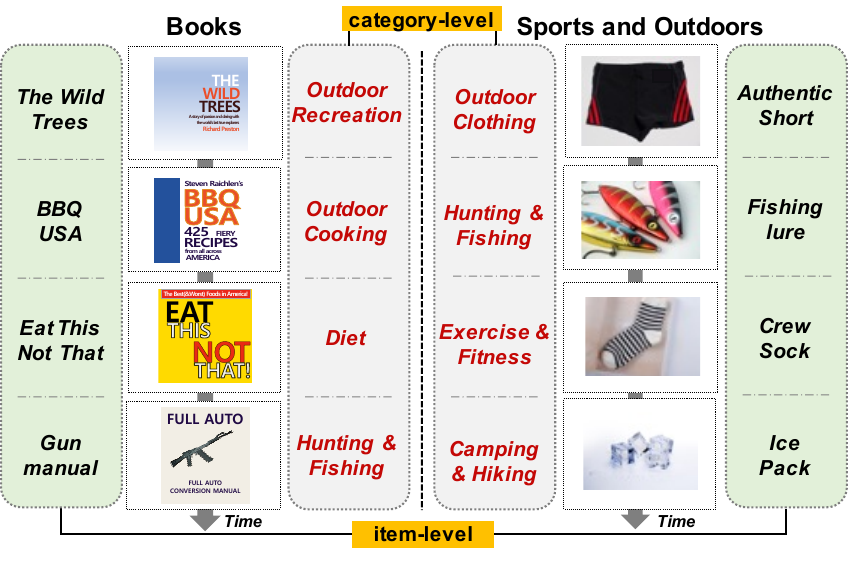}
    \end{center}
    \vspace{-0.3cm}
    \caption{Illustration of user interaction sequence for two domains in \textit{Amazon} dataset.
    The user sequences at the category level are more general and domain-relevant than those at the item level for the same user.}
    \label{fig:recgpt_category_example}
    \end{figure}

In addition, we develop a hierarchical contrastive learning method that sequentially learns user behavior patterns from the coarsest to the finest (i.e., item-level) categories of user interaction sequences, thereby discerning more general user preferences in each domain. 
These category-level sequences potentially provide more generalized user preferences and are more cross-domain relevant compared to the item-level preferences, as shown in Figure \ref{fig:recgpt_category_example}. 
These generalized user preferences play a pivotal role in diminishing the negative transfer effect caused by heavily biased user preferences in a specific domain. 
The proposed hierarchical contrastive learning method progresses from learning coarse-level categories down to fine-level ones, culminating at the item level.

As a result, our proposed CGRec allows the performance of the CDSR task to be improved with two real-world datasets for multiple domains. 
In particular, some of the domains that are performing poorly due to the negative transfer show large gains in performance in the model.

\vspace{-0.3cm}    
\section{Related Work} \label{section:related_work}
We focus on the CDSR task, which is a mixture of the CDR task for three or more domains and the SR task reflecting the temporal order of user sequences.
Since most of the CDSR studies are for two domains, our study is the first CDSR study for multiple domains to our knowledge.

\vspace{-0.15cm}
\subsection{Sequential Recommendation} \label{subsection:sr}
The sequential recommendation framework models the temporal dynamics of user-item interactions, by capturing user evolving preferences.
GRU4Rec \cite{hidasi2015session} is a session-based recommendation model using Gated Recurrent Units (GRU), capturing the complex dependencies within sessions. 
Also, attention mechanisms have been employed to identify and encapsulate sequential patterns, taking into account short- and long-term dependencies \cite{kang2018self,sun2019bert4rec,zhou2020s3,rashed2022context}. 
For instance, SASRec \cite{kang2018self} employed self-attention mechanisms to model user-item interactions while BERT4Rec \cite{sun2019bert4rec} integrated the \textit{Cloze} objective to construct a pre-trained model tailored for sequential recommendations. 
CORE \cite{hou2022core} unified the self-attention based encoding and decoding representation space by using a robust distance measure to prevent overfitting.

Meanwhile, there have been studies using item features together with item sequences \cite{zhang2019feature, zhou2020s3, rashed2022context}.
A feature-level self-attention layer was proposed to utilize the attribute information such as the category of the item in FDSA \cite{zhang2019feature}.
S$^{3}$Rec \cite{zhou2020s3}, in their pretraining phase, used four self-supervised objectives to train the relationships among attributes, items, subsequences, and sequences.
Adding extra information of items leads to significant performance improvement of S$^{3}$Rec, and this improvement is also observed in CARCA \cite{rashed2022context}. 
CARCA utilized not only categorical attributes, but also contextual information such as buying time and image of items. 
The dynamic nature of abundant context and attribute is captured by self-attention blocks, but CARCA uses encoder-decoder architectures instead of encoder-only or decoder-only model.
However, both models still have the negative transfer issues in the cross-domain recommendation.

\vspace{-0.3cm}
\subsection{Cross-Domain Recommendation}
The aim of Cross-Domain Recommendation (CDR) is to enhance the quality of recommendations in the target domain through the utilization of complementary knowledge from other domains.
For example, CMF \cite{singh2008relational} modeled the collective matrix factorization using the pairwise relational data while CLFM \cite{gao2013cross} introduced a cluster-level latent factor model, which captures domain-specific patterns and common patterns across domains.
Another approach, DTCDR \cite{zhu2019dtcdr}, provided user and item embeddings using rating and multi-source content, and applied a multi-task learning strategy to share these embeddings across domains, to improve the recommendation performance in both data-rich and sparse domains simultaneously.
BiTGCF \cite{liu2020cross} extracted user-item relationships in each domain and facilitated their transfer using the graph convolutional
networks.
DeepAPF \cite{yan2019deepapf} approximated complex user-video interactions, capturing cross-domain interests and single domain interests using an attention layer.
However, these methods approached the CDR task by modeling a pairwise domain-domain relationship, which loses effectiveness when dealing with a large number of domains. 
In order to reconcile this research deficit, CAT-ART \cite{li2023one} introduced a CDR method designed to enhance recommendations across $N\ge3$ participating domains. 
The method generated a global user representation and a domain-specific user embedding based on the matrix factorization. 
Yet, these studies overlooked the sequential characteristics inherent in user interactions.

\vspace{-0.25cm}
\subsection{Cross-Domain Sequential Recommendation}
The CDSR approach aims to improve the sequential recommendation task where items belong to multiple domains. 
This approach takes into account the sequential dynamics of user interactions. 
$\pi$-Net \cite{ma2019pi} devised gating mechanisms to transfer single-domain information to other paired domain. 
C$^{2}$DSR \cite{cao2022contrastive} modeled single- and cross-domain representation utilizing the self-attention based encoder and graph neural network with proposed cross-domain infomax objective.
However, these studies modeled a pairwise domain-domain relationship only. 
Applying the models to a situation involving $N\ge3$ domains would necessitate managing an excessive number of domain pairs, which may not be realistic in real-world applications \cite{li2023one}.
Therefore, we focus on extending CDSR to handle more than three domains as well as to consider the sequential dynamics, which has not been proposed in previous studies.

    \begin{figure*}[]
    \begin{center}
    \includegraphics[width=0.9\linewidth]{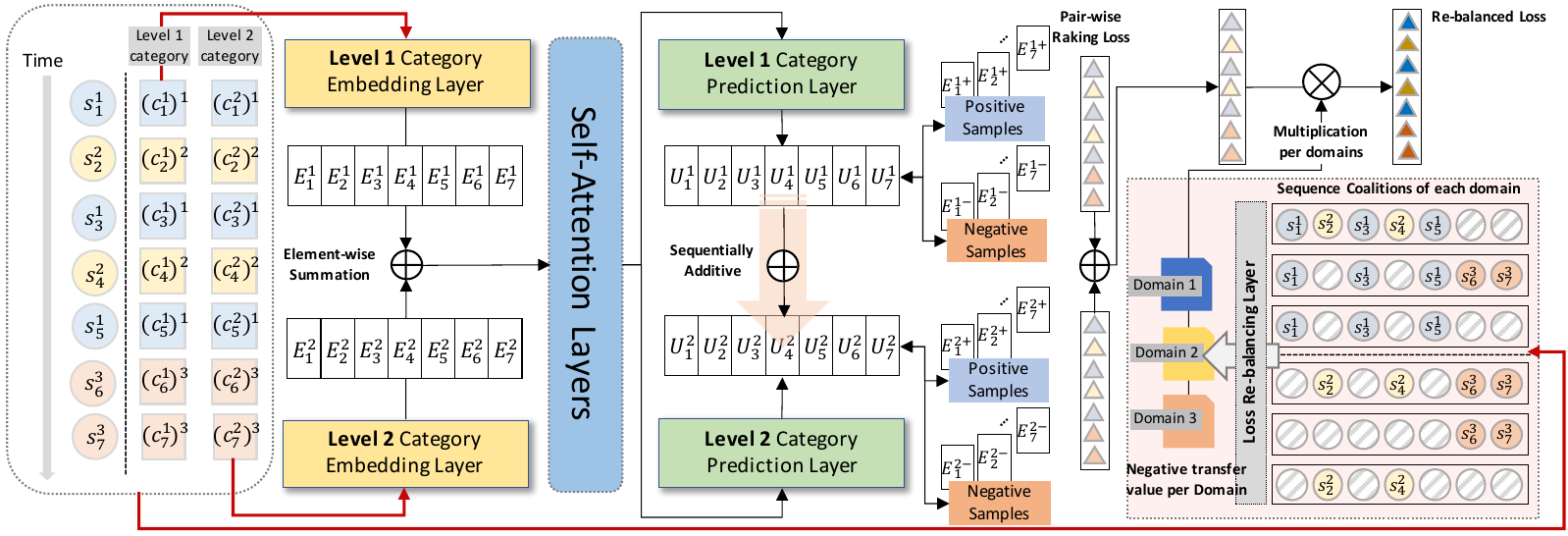}
    \end{center}
    \vspace{-0.3cm}
    \caption{We illustrate our model with three domains (represented by the colors blue, yellow, and pink in the left side of the figure) and two hierarchical categories.
    $\triangle$ indicates the loss per items.
    The domain hybrid sequence $S=[s^{1}_{1},s^{2}_{2},s^{1}_{3}, s^{2}_{4},s^{1}_{5}, s^{3}_{6},s^{3}_{7}]$ is input into our model. The first feeding process is expressed as the bold-red lines, and the sequence $S$ is represented as tuple of two categories sequences. 
    Note that the level 2 category indicates the item level in this figure.
    }
    \label{fig:overall_model}
    \end{figure*}

\vspace{-0.1cm}
\section{Preliminary} 
In this work, we focus on a expanding CDSR scenario, where each interaction sequence involves $D\ge3$ domains.
The domain $d\in[1,2,...,D]$ can be shopping categories, such as books and movies, or the different service platforms (e.g., short video platforms) \cite{zhu2021cross,li2023one}.

\noindent\textbf{Definition 1. Hierarchical Categories of Items}: 
Each item corresponds to various levels of categories.
We posit an $H$-level hierarchical categorization within each domain. 
In this hierarchy, level-$1$ signifies the coarsest-grained category, while level $H$ (i.e., item level), represents the finest category. 
Note that the item level is treated as the level $H$ category.
For instance, if the \textit{Books} domain is designated as the coarsest category level (level-$1$), the succeeding category level (level-$2$) may include various sub-categories such as \textit{Economics} or \textit{Psychology} in the \textit{Amazon} dataset.

Thus, we introduce the notation $I^{d,h}$ to denote a set of interactions within the level-$h$ category for domain $d$.
Additionally, the cumulative set of interactions across all category levels within domain $d$ is represented as $I^{d,\cdot}=\{I^{d,1},...,I^{d,H}\}$.
Moreover, the cumulative set of interactions can be defined across all domains within the level-$h$ category as $I^{\cdot,h}=\{I^{1,h},...,I^{D,h}\}$.

\noindent\textbf{Definition 2. Domain-hybrid Sequence}: 
The sequential interactions within the domain $d$ can be encapsulated as a sequence $S^d = \{s_{1}^{d}, s_{2}^{d}, ..., s_{|S^d|}^{d}\}$, where each $s_{t}^{d}\in S^{d}$ represents a tuple of interactions across all levels of categories $(c^{1}_{t},c^{2}_{t},...,c^{H}_{t})^d$ interacted with at time $t$.
Subsequently, we denote $S=(S^1, S^2,...,S^D)$ as the domain-hybrid sequence that has been shuffled according to chronological order.
As an illustration, it is assumed that a user interacted with three domains $[1,2,3]$, with the interaction sequences for each domain as $S^1=[s^{1}_{1},s^{1}_{2},s^{1}_{3}]$, $S^2=[s^{2}_{1},s^{2}_{2}]$, and $S^3=[s^{3}_{1},s^{3}_{2}]$. 
The sequence $S=[s^{1}_{1},s^{2}_{1},s^{1}_{2}, s^{2}_{2},s^{1}_{3}, s^{3}_{1},s^{3}_{2}]$ then symbolizes the domain-hybrid interaction sequence, which merges $S^1$, $S^2$, and $S^3$ in chronological order\footnote[1]{This sequence can be described as $[s^{1}_{1},s^{2}_{2},s^{1}_{3}, s^{2}_{4},s^{1}_{5}, s^{3}_{6},s^{3}_{7}]$.}. 
If domains 1 and 2 have two levels of categories, and domain 3 has three levels categories, then $s^{1}_{1}=(c^{1}_{1},c^{2}_{1})^1$, $s^{2}_{1}=(c^{1}_{1},c^{2}_{1})^2$, and $s^{3}_{2}=(c^{1}_{2},c^{2}_{2},c^{3}_{2})^3$.

\noindent\textbf{Problem Statement}: Given the observed domain-hybrid sequences until $t$ time step $S_{1:t}=(S^1, S^2,...,S^D)_{1:t}$, the goal of CDSR is to predict the tuple of next interactions $s^{n}_{t+1}=(c^{1}_{t+1},c^{2}_{t+1},...,c^{H}_{t+1})^n$:

 \begin{equation}
  \begin{split}
 \label{equation:problem_statement}
  &\argmax_{s^{n}_{t+1}\in I^{n,\cdot}} P(s^{n}_{t+1}|S_{1:t}), \;\; \text{if domain of the next item}=n,
 \end{split}
 \end{equation} 
where $P(s^{n}_{t+1}|S_{1:t})$ is the joint probability of the candidate interactions of all $H$ levels categories in domain $n$.
Consequently, our model is trained to predict the next interactions, considering all the categories at different hierarchical levels $(c^{1}_{t+1},...,c^{H}_{t+1})^n$ concurrently. 
In CDSR, the interaction from the last hierarchy level (i.e., item level; $c^{H}_{t+1}$) with the highest score is selected as the next recommended item.

\vspace{-0.2cm}
\section{Model}
\subsection{Item Embedding Layer} \label{subsection:item_embedding_layer}
The category sequence implies a more general user preference than the item sequence, which serves to mitigate negative transfers between irrelevant domains \cite{zhang2019feature}.
Derived from this hierarchy of items, we suggest considering the sequences of categories corresponding to these items as additional input features, and training them together with the proposed objective discussed in Section \ref{subsection: training_obj}.

First, each domain-hybrid sequence is converted to a fixed length ($=m$) sequence by truncating or padding interactions, as in previous studies \cite{kang2018self, zhou2020s3,cao2022contrastive}.
Then, the domain-hybrid sequence $S$ is transformed into a $m$-length embedding vector sequence.
The level-$h$ category embedding matrix of domain $d$, ${B}^{d,h}$, is represented by a matrix $\mathbb{R}^{|I^{d,h}| \times r}$, where $|I^{d,h}|$ is the size of the level-$h$ category sets in the domain $d$, and $r$ is the embedding dimension. 
Then, all embedding matrix of $D$ domains are concatenated to form the comprehensive level-$h$ category embedding matrix ${B}^{h}\in \mathbb{R}^{|I^{\cdot,h}| \times r}$.
Specifically, a look-up operation using ${B}^{h}$ was used in order to build up the input embedding matrix $\mathrm{E}^{h}\in \mathbb{R}^{m \times r}$. 
In addition, a trainable position embedding matrix $\mathrm{P}\in \mathbb{R}^{m \times r}$ is integrated to consider the sequential characteristic of the user interaction. 
Consequently, the final sequence representation $\mathrm{E}\in \mathbb{R}^{m \times r}$ is acquired by summing $H$+1 embedding matrices: $\mathrm{E}^{1}+\mathrm{E}^{2}+...+\mathrm{E}^{H}+P$. 
The embedding of the level-$h$ category in the $t$-th step is denoted as $\mathrm{E}^{h}_{t}$. 
This process results in an input sequence embedding for the self-attention encoder, which is described in the following section.

\vspace{-0.3cm}
\subsection{Self-Attention Encoder} \label{subsection:Encoder}
The self-attention module\cite{vaswani2017attention} is employed to encode the relationships between items within sequences.
This module typically incorporates two sub-layers as described in the following subsection. 

\vspace{-0.2cm}
\subsubsection{Multi-head Self-Attention}
This mechanism has shown its effectiveness in selectively garnering information from various representation subspaces.
The operation of this mechanism involves taking the embedding $\mathrm{E}$ as an input (refer to Section \ref{subsection:item_embedding_layer}), transforming $\mathrm{E}$ into three matrices via linear projections, and subsequently feeding the three matrices into an attention function. 
The multi-head self-attention (MHA) is defined as follows:

\vspace{-0.3cm}
 \begin{equation}
 \begin{split}
 \label{equation:multi_head}
  &MHA(Z^l)=[head_1 ^{\frown} head_2 ^{\frown}...^{\frown} head_p]\mathrm{W}^{F}, \\
  &head_i=Attention(Z^{l}\mathrm{W}^{Q}_{i},Z^{l}\mathrm{W}^{K}_{i},Z^{l}\mathrm{W}^{V}_{i}), 
 \end{split}
 \end{equation}
where the projection matrices $\mathrm{W}^{Q}_{i}$,$\mathrm{W}^{K}_{i}$,$\mathrm{W}^{V}_{i} \in \mathbb{R}^{r \times r/p}$ and $\mathrm{W}^{F} \in \mathbb{R}^{r \times r}$ are the trainable parameters, $Z^{l} \in \mathbb{R}^{|Z^l| \times d}$ is sequential input for the $l$-th self-attention layer, $p$ is the number of heads, and $^\frown$ is the concatenate operation.
When $l$=0, we set $Z^0=\mathrm{E}$.
$Z_{t}^{l}$ is the $t$-th step embedding of the $Z^{l}$. 
The scaled dot-product operation adopts the attention mechanism as follows:
\begin{equation}
 \label{equation:self_attention}
  Attention(Q,K,V)=softmax(\frac{QK^{T}}{\sqrt{r/p}})V,
 \end{equation}
where $Q=Z^{l}\mathrm{W}_{i}^{Q}$, $K=Z^{l}\mathrm{W}_{i}^{K}$, and $V=Z^{l}\mathrm{W}_{i}^{V}$ denote the output of linear projections of the embedding sequence, while $r/p$ serves as a scale factor. 
In our model, $L$ multi-head self-attention layers are stacked in the self-attention encoder.

\subsubsection{Point-Wise Feed-Forward Network}
We adopt a point-wise feed-forward network to all $Z_t^l$ to incorporate nonlinearity into the model and facilitate interactions among different latent subspaces as follows:

\vspace{-0.3cm}
 \begin{equation}
 \begin{split}
 \label{equation:ffn}
  &Z^l=[FFN(Z_1^l)^{\top \frown} FFN(Z_2^l)^{\top \frown} \cdots{^\frown} FFN(Z_m^l)^{\top}], \\
  &FFN(Z_t)=GELU(Z_{t}\mathrm{W}_{1}+b_{1})\mathrm{W}_{2}+b_{2},
 \end{split}
 \end{equation}
where $\mathrm{W}_1$, $b_1$,$\mathrm{W}_2$, and $b_2$ represent trainable parameters, $GELU$ is the gelu activation \cite{hendrycks2016gaussian}, and $^\frown$ symbolizes the concatenation.
In the sequential recommendation task, only use the information prior to the current time step can be used.
Therefore, a masking operation is employed in the output sequences of the multi-head self-attention layer to remove all links between $Q_i$ and $K_j$ when $j>i$. 

\vspace{-0.19cm}
\subsection{Hierarchical Prediction Layer} \label{subsection:Hierarchical Prediction Layer}
In our model, sequences of categories from various hierarchies ($H$) are trained simultaneously.
During the training process, the interdependencies among these sequences of categories are integrated by transferring information from coarse-grained to fine-grained categories. 
This strategy ensures an unbiased representation at the finest category level, i.e., the item level, by incorporating the patterns of coarser categories that reflect broader user preferences.
The purpose of this approach is to prevent the negative transfer issue - a scenario in which a user preference in one domain excessively impacts other unrelated domains, subsequently reducing the overall performance. 

Upon applying $L$ multi-head self-attention layers, the next interaction is predicted using the final representation from the self-attention encoder given the first $t$ items.
Specifically, the user's preference score is initially computed  for the level-$1$ category (i.e., the coarsest category) in step $t+1$, given the user's historical context as follows:

\vspace{-0.3cm}
 \begin{equation}
 \label{equation:prediciton_category}
 \begin{split}
  &P(c_{t+1}^{1}=c^{1}|S_{1:t})=e_{c^1}^{\top}\cdot U_{t}^{L,1}, \\
  &U_{T}^{L,1}=FFN^{1}(Z_{t}^{L}), \\
  &FFN^{1}(Z_{t})=GELU(Z_{t}\mathrm{W}_{3}^{1}+b_{3}^{1})\mathrm{W}_{4}^{1}+b_{4}^{1},
  \end{split}
 \end{equation}
where $e_{c^1}$ is the level-$1$ representation of positive item $c^{1}$ from the embedding matrix $\mathrm{E}^{1}$, and $U_{t}^{L,1}$ is the output of feed-forward network for level-$1$ prediction.
In addition, $Z_{t}^{L}$ is the output of the $L$-th self-attention layer at step $t$, and $\mathrm{W}_{3}^{1}$, $b_{3}^{1}$,$\mathrm{W}_{4}^{1}$, $b_{4}^{1}$ are trainable parameters.
Then, the preference score for the succeeding level $2$ category in the step $t+1$ is sequentially calculated as:
 \begin{equation}
 \label{equation:prediciton_category}
  P(c_{t+1}^{2}=c^{2}|S_{1:t})=e_{c^{2}}^{\top}\cdot (U_{t}^{L,2}\oplus U_{t}^{L,1}),
 \end{equation}
where $e_{c^{2}}$ is the level-$2$ embedding of positive item $c^{2}$ from category embedding matrix $\mathrm{E}^{2}$ and $\oplus$ indicates the element-wise sum operation.
This operation incorporates information from the coarse-grained categories into the fine-grained categories when calculating the preference score, and therefore the predictions of fine-grained categories depend on the predicted outcomes of subsequent coarse-grained categories.
This interdependence between hierarchical levels highlights the significance of integrating information across multiple categories to gain a comprehensive understanding of the user preferences.
From this process, the preference score for the level-$H$ category (i.e., item level) in the step $t+1$ is extracted as follows:
 \begin{equation}
 \label{equation:prediciton_item}
  P(c_{t+1}^{H}=c^{H}|S_{1:t})=e_{c^H}^{\top}\cdot (U_{t}^{L,H}\oplus U_{t}^{L,(H-1)}).
 \end{equation}

\subsection{Training Objective} \label{subsection: training_obj}
Previous studies \cite{zhang2019feature, rashed2022context} have used these categories as auxiliary inputs but have overlooked the interdependency between items and their categories during the training process. 
We therefore adopt a hierarchical approach to contrastive learning, first predicting the coarsest categories, and then using the learned representations to predict finer-grained categories. 

To simultaneously maximize all preference scores across all category levels and domains, the pairwise rank loss $\mathcal{L}^{hcross}$ is incorporated as follows:
 \begin{equation}
 \label{equation:hierarchical_constrastive_learning}
 \begin{split}
  &\mathcal{L}^{hcross}=\sum_{h=1}^{H}\sum_{t=1}^{m}\mathrm{log}\sigma\Bigg(P(c_{t+1}^{h}=c^{h}|S_{1:t})  -P(c_{t+1}^{h}=c^{h-}|S_{1:t})\Bigg),
 \end{split}
 \end{equation}
where $\sigma$ denotes the sigmoid function, and the ground-truth item for the level-$h$ category $c^{h}$ is associated with a negative item $c^{h-}$ that is chosen through random sampling.

    \begin{figure}[]
    \begin{center}
    \includegraphics[width=1 \linewidth]{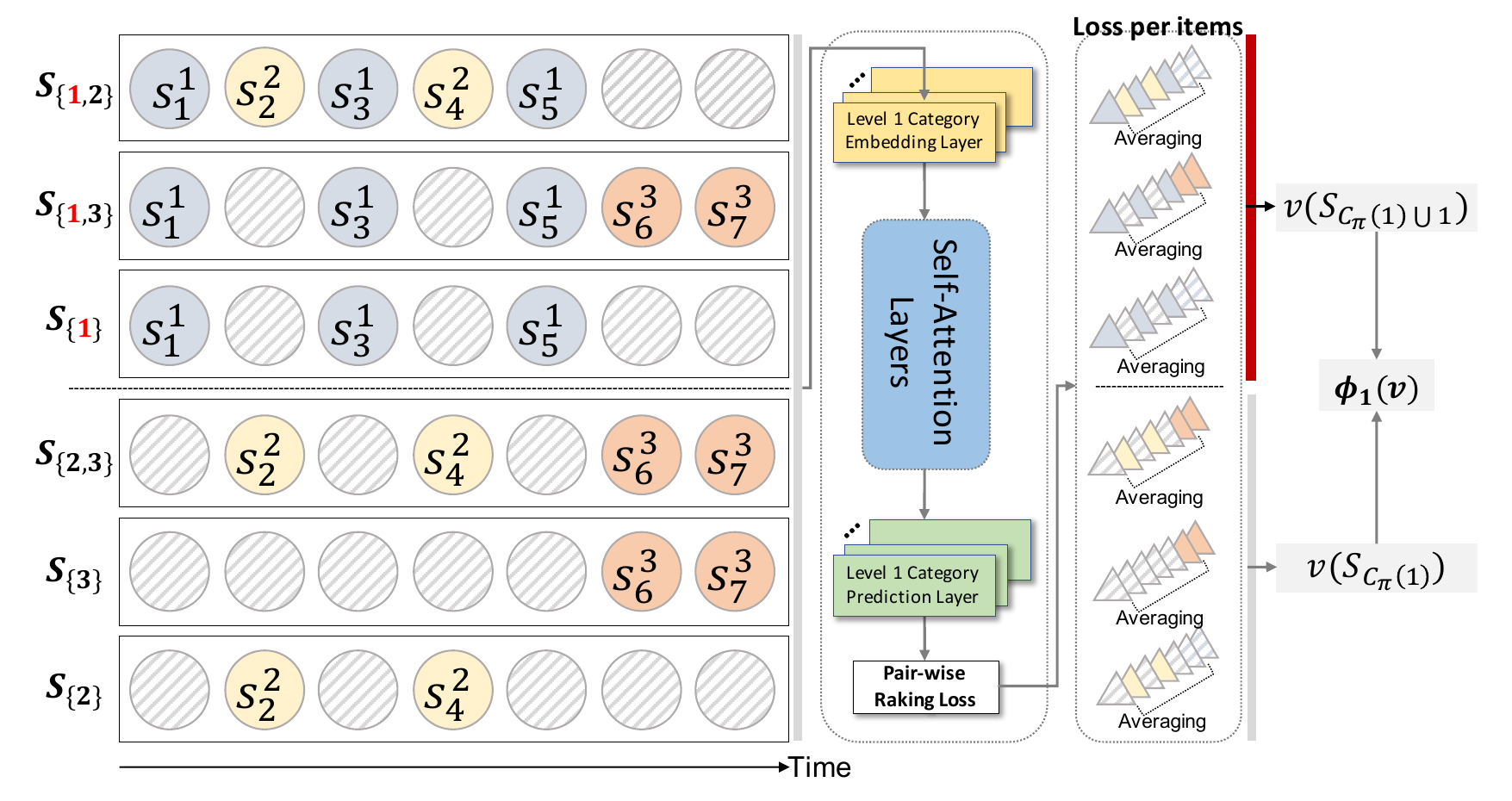}
    \end{center}
    \vspace{-0.3cm}
    \caption{The process of obtaining the negative transfer value for the domain 1. 
    }
    \label{fig:shapley_example}
    \end{figure}

\vspace{-0.1cm}
\subsection{Loss Re-balancing Layer}
We focus on quantifying the negative transfer of a specific domain by measuring its marginal contribution for the training loss.
We consider negative transfer and marginal contribution to be opposite concepts.
Therefore, the larger the marginal contribution, the smaller the degree of negative transfer.
Low weight values are assigned for interacted items of high negative transfer (i.e., low marginal contribution) domains in the training process.
Our model was trained using the domain hybrid sequences, which considers each domain as players of a cooperative game \cite{myerson1991game}.
Then, Shapley values \cite{straffin1993game} were derived for all domains and were used to approximate the degree of the negative transfer of each domain.
Before exploring the solution methodology, it is pertinent to briefly discuss some foundational concepts from the cooperative game theory.

\subsubsection{Cooperative Games}
In the realm of game theory, a cooperative game \cite{myerson1991game} is distinguished by the emphasis on cooperative behavior among clusters of players, referred to as coalitions. 
More formally, a cooperative game is often represented by a pair $(N, v)$, where $N$ is a finite set of players denoted by $N = \{1, 2, ..., n\}$, and $v$ is a characteristic function. This characteristic function $v: 2^N \rightarrow \mathbb{R}$ allocates to each coalition $S \subseteq N$ a real number $v(S)$, signifying the maximum cumulative payoff that players in $S$ can attain through cooperation. For any subset $S \subseteq N$, $v(S)$ quantifies the worth of coalition $S$. By default, we set $v(\emptyset) = 0$ and $v(N) = 0$.

\subsubsection{Shapley Value}
The Shapley value \cite{straffin1993game} offers an approach to equitably distribute the cumulative payoff of the grand coalition, i.e., the coalition that includes all players within the game.
Shapley suggested a method that evaluates each player's role within the game by assessing its marginal contributions across all potential coalitions to which it could be a member.
Formally,  $\pi \in \Pi(N)$ is denoted as a permutation of players in $N$, and $C_{\pi}(i)$ as any coalition that exclude player $i$.
Then the Shapley value of player $i$, $\phi_{i}(v)$, is defined as follows:
 \begin{equation}
 \label{equation:shapley_value}
  \phi_{i}(v)=\frac{1}{|N|!}\sum_{\pi \in \Pi(N)}[v(C_{\pi}(i) \cup \{i\}) - v(C_{\pi}(i))],
 \end{equation}
where $v$ is the characteristic function discussed in the next section.
This value is the average marginal contribution of the player $i$.

\subsubsection{Quantifying negative transfer of each Domain}
In our model, each player $i$ is a specific domain $d$ in a sequence.
$S_{\pi}$ is defined as a sequence coalitions.
As shown in Figure \ref{fig:shapley_example}, a specific sequence is assumed to have three domains with seven interactions, which indicates $S=[s^{1}_{1},s^{2}_{2},s^{1}_{3}, s^{2}_{4},s^{1}_{5}, s^{3}_{6},s^{3}_{7}]$.
Then $2^{3}=8$ coalitions can be derived based on the combinations of three domains, and one example sequence with domains $2$ and $3$ is $S_{\{2,3\}}=[(pad),s^{2}_{2},(pad), s^{2}_{4},(pad), \\ s^{3}_{6},s^{3}_{7}]$.
After, the degree of the negative transfer of each domain is computed by making use of the following cooperative game scheme.
Note that the negative transfer and marginal contribution are opposite concepts.
We define a cooperative game, $(D,v)$, wherein the set of domains is the set of players and $v$ is a characteristic function that attaches a value for each domain subset $\pi \in \Pi(D)$ in a specific sequence.
In the example above, the domain hybrid sequence $S=[s^{1}_{1},s^{2}_{2},s^{1}_{3}, s^{2}_{4},s^{1}_{5}, s^{3}_{6},s^{3}_{7}]$ has 8 sequence coalitions, i.e., $\pi \in \{\emptyset,\{1\},\{2\},\{3\},\{1,2\},\{1,3\},\{2,3\},$
$\{1,2,3\}\}$. 
The value $v$ of each $\pi \in \Pi(D)$ is defined as the training loss (See Section \ref{subsection: training_obj}) of the model built using updates only from sequence  $S_{\pi}$:
 \begin{equation}
 \label{equation:shapley_value_2}
 \begin{split}
 &v(S_{\pi})=\sum_{h=1}^{H}\sum_{t\in (S_{\pi})_{T}}\mathrm{log}\sigma\Bigg(P(c_{t+1}^{h}=c^{h}|S_{1:t})-P(c_{t+1}^{h}=c^{h-}|S_{1:t})\Bigg) \\
 &\phi_{i}(v)=\frac{1}{|D|!}\sum_{\pi \in \Pi(N)}[v(S_{C_{\pi}(i) \cup \{i\}}) - v(S_{C_{\pi}(i)})],
 \end{split}
 \end{equation}
where $\phi_{i}(v)$ is the marginal contribution of the domain $i$, and $(S_{\pi})_{T}$ is the set of time steps of $S_{\pi}$. 
Note that this module shares the embedding layers (Section \ref{subsection:item_embedding_layer}), self-attention encoder (Section \ref{subsection:Encoder}), and hierarchical prediction layers (Section \ref{subsection:Hierarchical Prediction Layer}) to calculate $P(c_{t+1}^{h}=c^{h}|S_{1:t})-P(c_{t+1}^{h}=c^{h-}|S_{1:t})$.

Then, the relative value of the negative transfer in each domain is calculated.
Let $\gamma=(\gamma_{1},\gamma_{2},...,\gamma_{D})$ be the vector of the degree of the negative transfer in each domain wherein $\gamma_{d}$ represents the quantity of the negative transfer of domain $d$.
In this case, $\gamma_{d}$ is called \textit{negative transfer value} of domain $d$.
We posit that the more the negative transfer value for a domain, the less its contribution to the objective of our model.
The negative transfer vector is initialized to be uniform $\gamma_{d}=1/D$.
Then at each mini-batch in the training, the Shapley value is obtained for all domains from the cooperative game $(D,v)$ mentioned above.
The computed Shapley values is denoted as $\phi_{d}(v)$ for each $d\in D$
and the negative transfer vector $\gamma$ is updated as follows:
 \begin{equation}
 \label{equation:negative_transfer_vector}
 \gamma_{d} \leftarrow \alpha * \gamma_{d} +\beta * \phi_{d}(v); \forall d \in D,
 \end{equation}
where $\alpha$ and $\beta$ are trainable parameters.
The relative negative transfer value of each domain is determined by marginalizing $\gamma$ as $\gamma \leftarrow \texttt{softmax}(\gamma;\lambda)$, where $\texttt{softmax}$$(;\lambda)$ is the softmax function with the temperature $\lambda$ \cite{szegedy2016rethinking}.
When $\lambda \rightarrow \infty$, then $\gamma_{d}$ is uniform ($=1/D$).
Shapley values are typically approximated using Monte Carlo simulations due to their exponential computational complexity, which grows with the number of players involved. 
In all of our experimental settings, there are at most five domains participating in the game. 
With this number of players, calculating exact Shapley values involves computations over $2^{5}=32$ permutations, which is considered manageable. 
Therefore, Monte Carlo simulation was not applied to the Shapley value calculation in this paper. 

\subsubsection{Loss Re-balancing}
The relative negative transfer value $\gamma$ is used as a weight for the loss calculated in Eq. \ref{equation:hierarchical_constrastive_learning}.
A domain hybrid sequence comprises items ranging from 1 to a fixed length $m$, each of which belongs to a different domain. 
Consequently, the loss in Eq. \ref{equation:hierarchical_constrastive_learning} is aggregated separately depending on the item's domain, and the relative negative transfer value $\gamma_{d}$ is multiplied by the domain as follows:

\vspace{-0.2cm}
 \begin{equation}
 \label{equation:loss_rebalancing}
 \begin{split}
  &\mathcal{L}^{hcross'}=\sum_{h=1}^{H}\sum_{t=1}^{m}\sum_{d=1}^{D} \gamma_{d} \mathrm{log}\sigma\Bigg(P(c_{t+1}^{h}=c^{h}|S_{1:t})  -P(c_{t+1}^{h}=c^{h-}|S_{1:t})\Bigg).
 \end{split}
 \end{equation}
Then, the re-balanced loss $\mathcal{L}^{hcross'}$ is minimized with the gradient descent algorithm.

\section{Experiments}
The experiments aim to address the following research questions:

\noindent\textbf{(RQ1)}: Can our model surpass the performance of state-of-the-art CD(S)R and SR models that involve multiple domains ($D\ge3$)?

\noindent\textbf{(RQ2)}: Is our model capable of mitigating the issue of the negative transfer in the context of the CDSR task?

\noindent\textbf{(RQ3)}: How do the components of our model affect performance on the CDSR tasks?

\vspace{-0.2cm}
\subsection{Datasets}
Table \ref{tab:data_summary} describes the details of datasets used in this study.

\begin{table}[] 
\caption{Statistics of datasets}\label{tab:data_summary}
\vspace{-0.3cm}
\begin{adjustbox}{width=\columnwidth,center}
\begin{tabular}{c|ccccc}
\hline\hline
\textbf{Dataset}                 & \textbf{Domain}     & \textbf{\#Users}                & \textbf{\#Items} & \textbf{\#Interactions} & \textbf{\#Sparsity} \\ \hline
\multirow{5}{*}{Amazon} & Books      & \multirow{5}{*}{34,771} & 87,758   & 582,541         & 99.98\%    \\
                        & Clothing   &                        & 69,480   & 370,169         & 99.98\%    \\
                        & Movies     &                        & 36,070   & 343,056         & 99.97\%    \\
                        & Toys       &                        & 47,940   & 279,011         & 99.98\%    \\
                        & Sports     &                        & 57,387   & 242,169         & 99.99\%    \\ \hline
\multirow{5}{*}{T-Seq}  & App-Use    & \multirow{5}{*}{99,936} & 9,192    & 36,716,069       & 96.00\%    \\
                        & Call-Use   &                        & 3,301    & 3,038,385        & 99.07\%    \\
                        & PoI        &                        & 549     & 1,892,868        & 96.33\%    \\
                        & E-comm     &                        & 714     & 230,233         & 99.55\%    \\
                        & Membership &                        & 72      & 401,065         & 91.97\%    \\ \hline\hline
\end{tabular}
\end{adjustbox}
\end{table}

\vspace{-0.1cm}
\subsubsection{\textbf{Amazon}} 
We obtained Amazon review datasets \cite{mcauley2015image} with five domains, i.e., \textit{Books}, \textit{Clothing Shoes and Jewelry}, \textit{Movies and TV}, \textit{Toys and Games}, and \textit{Sports and Outdoors}.
They are represented by the abbreviations \textit{Books}, \textit{Clothing}, \textit{Movies}, \textit{Toys}, and \textit{Sports} respectively.
Users who have interactions across five domains were identified, and then retained domain-hybrid sequences that contain at least two items from each domain.

\vspace{-0.1cm}
\subsubsection{\textbf{T-Seq}} 
User logs were collected from five domains using various real-world applications owned by a global telecommunications company. 
The dataset, which included  99,936 customers who had agreed to have their information be collected and analyzed, was randomly sampled.
The information included Application usage (\textit{APP-Use}), Call detailed record (\textit{Call-Use}), PoI preference (\textit{PoI}), e-commerce purchase (\textit{e-comm}), and membership usage (\textit{MBR}).

\vspace{-0.1cm}
\subsection{Experimental Setup}
\subsubsection{Evaluation Metrics}
Consistent with previous studies \cite{kang2018self, zhou2020s3, rashed2022context}, the \textit{leave-one-out} method for measuring recommendation performance was used. 
In particular, every sequence of user interactions is segmented into three segments: the last item is designated as test data, the item preceding the last item is set aside for validation data, and all remaining data is used for training purposes.
The performance of each domain was reported separately based on the domain of the last item in the test data.
Therefore, performance per domain was measured with different users.  
Given the expansive item set, using all items as testing candidates would be prohibitively time-consuming. 
As a result, a widely accepted strategy that pair the ground truth item (i.e., positive sample) with 99 randomly selected items (i.e., negative samples) that the user had not previously interacted with was employed.
Evaluation metrics were then extracted based on the ranking of these 100 items, and an average score across all users is presented. 
Following this, the Top-$k$ recommendation performance is reported, derived from a ranked list of 100 items. 
Performance was assessed in terms of Hit Ratio (HR@\{5, 10\}), Normalized Discounted Cumulative Gain (NDCG@\{5, 10\}), and Mean Reciprocal Rank (MRR).

\vspace{-0.1cm}
\subsubsection{Compared Baselines}
We compared the performance of our model to three categories of baselines: (1) traditional recommendation (TR), (2) sequential recommendation (SR), and (3) cross-domain (sequential) recommendation (CD(S)R) as shown in Table \ref{tab:result_1}. 
Traditional recommendation includes BPRMF and FPMC. 
BPRMF \cite{rendle2012bpr} is a classic method, designed for developing recommendations from implicit feedback data, that introduces a pairwise loss function to model users' relative preferences.
FPMC \cite{rendle2010factorizing} uses personalized transition graphs over the markov chains to learn unique user transition matrices.
See Section \ref{section:related_work} for a description of SR and CD(S)R baselines.
All the CDR baselines and some SR models were implemented using the RecBole framework \cite{zhao2021recbole,recbole[2.0]}.
CARCA, S$^{3}$Rec, C$^{2}$DSR, and CAT-ART were implemented using their official code.

\vspace{-0.1cm}
\subsubsection{Hyperparameter Setting.}
For all models, certain hyperparameters were set. 
The embedding size $r$ and mini-batch size was set to 128, and the dropout rate was fixed at 0.2. 
The training epoch was set at 100 to achieve optimal results and the Adam optimizer was employed to update all parameters. 
The best evaluation results were chosen based on the highest NDCG@5 performance on the validation set. 
In our model, we applied the number of category levels as two and three in \textit{Amazon} and \textit{T-Seq}, respectively.
Moreover, we utilized two single-head attention blocks with four head size.

\begin{table*}  \footnotesize
\centering
\caption{Comparison of the Proposed Method and Baselines. The top two methods are highlighted in bold and underlined.}
    \vspace{-0.35cm}
    \setlength{\tabcolsep}{7.5pt}
    \renewcommand{\arraystretch}{0.87}
\label{tab:result_1}
\begin{tabular}{c|c|c|ccccc|ccccc} 
\hline\hline
\multicolumn{1}{c|}{\multirow{2}{*}{\textbf{Methods}}} & \multirow{2}{*}{\textbf{Models}}    & \textbf{Dataset} & \multicolumn{5}{c|}{\textbf{Amazon}}                                                    & \multicolumn{5}{c}{\textbf{T-Seq}}                                                       \\ 
\cline{3-13}
\multicolumn{1}{c|}{}                         &                                      & Domains          & Book            & Clothing        & Movie           & Toys            & Sports          & App-Use         & Call-Use        & PoI             & MBR             & e-comm           \\ 
\hline
\multicolumn{1}{c|}{\multirow{10}{*}{\textbf{TR}}}     & \multirow{5}{*}{\textbf{BPRMF}}      & HR@5             & 0.2382          & 0.2377          & 0.2402          & 0.2386          & 0.2398          & 0.9394          & 0.6387          & 0.8516          & 0.8657          & 0.1075           \\
\multicolumn{1}{c|}{}                         &                                      & HR@10            & 0.3458          & 0.3454          & 0.3468          & 0.3466          & 0.3445          & 0.9697          & 0.7993          & 0.9422          & 0.9281          & 0.3306           \\
\multicolumn{1}{c|}{}                         &                                      & NDCG@5           & 0.1628          & 0.1630          & 0.1642          & 0.1634          & 0.1635          & 0.8578          & 0.4488          & 0.6589          & 0.6880          & 0.049            \\
\multicolumn{1}{c|}{}                         &                                      & NDCG@10          & 0.1976          & 0.1977          & 0.1985          & 0.1981          & 0.1972          & 0.8677          & 0.5010          & 0.6886          & 0.7084          & 0.1204           \\
\multicolumn{1}{c|}{}                         &                                      & MRR              & 0.1524          & 0.1527          & 0.1534          & 0.1529          & 0.1523          & 0.8343          & 0.4075          & 0.6069          & 0.6363          & 0.0595           \\ 
\cline{2-13}
\multicolumn{1}{c|}{}                         & \multirow{5}{*}{\textbf{FPMC}}       & HR@5             & 0.2763          & 0.1848          & \textbf{0.5994} & 0.2310          & 0.1926          & 0.9418          & 0.6728          & \underline{0.8996}  & 0.8689          & 0.3319           \\
\multicolumn{1}{c|}{}                         &                                      & HR@10            & 0.3939          & 0.2762          & \textbf{0.6782} & 0.3414          & 0.2853          & 0.9709          & 0.8101          & 0.9608          & 0.9254          & 0.4606           \\
\multicolumn{1}{c|}{}                         &                                      & NDCG@5           & 0.1932          & 0.1253          & \textbf{0.4910} & 0.1595          & 0.1302          & 0.8521          & 0.5265          & 0.7487          & 0.7169          & 0.2492           \\
\multicolumn{1}{c|}{}                         &                                      & NDCG@10          & 0.2309          & 0.1547          & \textbf{0.5164} & 0.1950          & 0.1600          & 0.8616          & 0.5712          & 0.7687          & 0.7355          & 0.2902           \\
\multicolumn{1}{c|}{}                         &                                      & MRR              & 0.1813          & 0.1179          & \textbf{0.4654} & 0.1506          & 0.122           & 0.8257          & 0.4967          & 0.7065          & 0.6738          & 0.2384           \\ 
\hline\hline
\multirow{35}{*}{\textbf{SR}}                          & \multirow{5}{*}{\textbf{GRU4Rec}}    & HR@5             & 0.2899          & 0.2861          & 0.5281          & 0.2929          & 0.2769          & 0.9504          & 0.7016          & 0.9214          & 0.8807          & 0.3885           \\
                                              &                                      & HR@10            & 0.4129          & 0.3977          & 0.6235          & 0.4032          & 0.3843          & 0.9774          & 0.8496          & \underline{0.9716}  & 0.9495          & 0.5355           \\
                                              &                                      & NDCG@5           & 0.1980          & 0.1963          & 0.4131          & 0.2009          & 0.1925          & 0.8532          & 0.5172          & 0.7659          & 0.7029          & 0.2879           \\
                                              &                                      & NDCG@10          & 0.2375          & 0.2322          & 0.4440          & 0.2364          & 0.2272          & 0.8621          & 0.5654          & 0.7824          & 0.7252          & 0.3343           \\
                                              &                                      & MRR              & 0.1840          & 0.1816          & 0.3877          & 0.1853          & 0.1791          & 0.8239          & 0.4762          & 0.7205          & 0.6522          & 0.2733           \\ 
\cline{2-13}
                                              & \multirow{5}{*}{\textbf{SASRec}}     & HR@5             & 0.2588          & 0.2623          & 0.4781          & 0.2724          & 0.2473          & 0.9331          & 0.6677          & 0.8961          & 0.8576          & 0.3333           \\
                                              &                                      & HR@10            & 0.3659          & 0.3701          & 0.5730          & 0.3807          & 0.3520          & 0.9656          & 0.8154          & 0.9636          & 0.9497          & 0.4530           \\
                                              &                                      & NDCG@5           & 0.1819          & 0.1888          & 0.3813          & 0.1967          & 0.1761          & 0.8400          & 0.5107          & 0.7537          & 0.6528          & 0.2622           \\
                                              &                                      & NDCG@10          & 0.2163          & 0.2235          & 0.4118          & 0.2315          & 0.2098          & 0.8506          & 0.5585          & 0.7759          & 0.6831          & 0.3004           \\
                                              &                                      & MRR              & 0.191           & 0.1989          & 0.3768          & 0.2062          & 0.1864          & 0.8148          & 0.4883          & 0.7178          & 0.6002          & 0.2785           \\ 
\cline{2-13}
                                              & \multirow{5}{*}{\textbf{FDSA}}       & HR@5             & 0.3040          & \underline{0.3159}  & 0.4381          & \underline{0.3002}  & \underline{0.3012}  & 0.9360          & 0.6622          & 0.8900          & 0.8627          & 0.3066           \\
                                              &                                      & HR@10            & 0.4218          & \textbf{0.4406} & 0.5465          & 0.4146          & \underline{0.4092}  & 0.9719          & 0.8155          & 0.9686          & 0.9318          & 0.4569           \\
                                              &                                      & NDCG@5           & 0.2166          & \underline{0.2210}  & 0.3403          & 0.2105          & \underline{0.2118}  & 0.8246          & 0.4645          & 0.7119          & 0.6551          & 0.2187           \\
                                              &                                      & NDCG@10          & 0.2544          & \underline{0.2614}  & 0.3754          & 0.2474          & \underline{0.2466}  & 0.8364          & 0.5147          & 0.7378          & 0.6776          & 0.2663           \\
                                              &                                      & MRR              & 0.2033          & 0.2065          & 0.3224          & 0.1961          & 0.1967          & 0.7918          & 0.4199          & 0.6632          & 0.5947          & 0.2088           \\ 
\cline{2-13}
                                              & \multirow{5}{*}{\textbf{BERT4Rec}}   & HR@5             & 0.2852          & 0.2683          & 0.4993          & 0.2772          & 0.2299          & 0.9385          & 0.6544          & 0.8404          & 0.8357          & 0.2218           \\
                                              &                                      & HR@10            & 0.3900          & 0.3798          & 0.6042          & 0.3934          & 0.3313          & 0.9704          & 0.8040          & 0.9349          & 0.9281          & 0.4178           \\
                                              &                                      & NDCG@5           & 0.2031          & 0.1888          & 0.3950          & 0.1923          & 0.1583          & 0.8480          & 0.4740          & 0.6567          & 0.6406          & 0.1315           \\
                                              &                                      & NDCG@10          & 0.2369          & 0.2247          & 0.4288          & 0.2299          & 0.1909          & 0.8585          & 0.5225          & 0.6877          & 0.6708          & 0.1943           \\
                                              &                                      & MRR              & 0.1900          & 0.1775          & 0.3744          & 0.1800          & 0.1482          & 0.8217          & 0.4345          & 0.6084          & 0.5878          & 0.1277           \\ 
\cline{2-13}
                                              & \multirow{5}{*}{\textbf{S$^{3}$Rec}}      & HR@5             & 0.3205          & 0.3086          & 0.3215          & 0.2989          & 0.2657          & 0.9465          & \underline{0.7615}  & 0.9289          & \underline{0.9240}  & \underline{0.4603}   \\
                                              &                                      & HR@10            & 0.4253          & 0.4189          & 0.4278          & 0.4076          & 0.3616          & 0.9734          & \underline{0.8776}  & \textbf{0.9787} & \textbf{0.9751} & \underline{0.5757}   \\
                                              &                                      & NDCG@5           & \underline{0.2377}  & 0.228           & 0.2503          & 0.2158          & 0.1966          & 0.8814          & \underline{0.6024}  & \textbf{0.7899} & \underline{0.7842}  & 0.2964           \\
                                              &                                      & NDCG@10          & 0.2715          & 0.2636          & 0.2846          & 0.2507          & 0.2275          & 0.8891          & \underline{0.6405}  & \underline{0.8030}  & \underline{0.7983}  & \underline{0.3509}   \\
                                              &                                      & MRR              & 0.2432          & \underline{0.2359}  & 0.2591          & \underline{0.2340}  & 0.2058          & 0.8602          & \underline{0.5721}  & \textbf{0.7829} & \textbf{0.7676} & \underline{0.3031}   \\ 
\cline{2-13}
                                              & \multirow{5}{*}{\textbf{CORE}}       & HR@5             & 0.2673          & 0.1949          & 0.4334          & 0.2331          & 0.2003          & 0.9382          & 0.7002          & 0.8527          & 0.8408          & 0.3242           \\
                                              &                                      & HR@10            & 0.3537          & 0.2699          & 0.5123          & 0.3111          & 0.2795          & 0.9712          & 0.8122          & 0.9406          & 0.9254          & 0.4451           \\
                                              &                                      & NDCG@5           & 0.2004          & 0.1491          & 0.3572          & 0.1753          & 0.1515          & 0.8398          & 0.5727          & 0.6886          & 0.6683          & 0.2550           \\
                                              &                                      & NDCG@10          & 0.2282          & 0.1731          & 0.3826          & 0.2003          & 0.1769          & 0.8506          & 0.6090          & 0.7174          & 0.6963          & 0.2935           \\
                                              &                                      & MRR              & 0.1898          & 0.1439          & 0.3425          & 0.1665          & 0.1438          & 0.8110          & 0.5456          & 0.6460          & 0.6228          & 0.2476           \\ 
\cline{2-13}
                                              & \multirow{5}{*}{\textbf{CARCA}}      & HR@5             & 0.2931          & 0.2716          & 0.4017          & 0.2770          & 0.2662          & 0.8582          & 0.6192          & 0.8281          & 0.8722          & 0.2510           \\
                                              &                                      & HR@10            & 0.3977          & 0.4112          & 0.5709          & 0.3663          & 0.3775          & 0.9052          & 0.7593          & 0.9142          & 0.9130          & 0.4943           \\
                                              &                                      & NDCG@5           & 0.2065          & 0.1844          & 0.3702          & 0.2027          & 0.1835          & 0.7598          & 0.4326          & 0.6411          & 0.6782          & 0.1347           \\
                                              &                                      & NDCG@10          & 0.2398          & 0.2293          & 0.4008          & 0.2313          & 0.2193          & 0.7751          & 0.4783          & 0.6696          & 0.6918          & 0.2133           \\
                                              &                                      & MRR              & 0.2093          & 0.1970          & 0.3592          & 0.2117          & 0.1919          & 0.7370          & 0.3994          & 0.5963          & 0.6201          & 0.1498           \\ 
\hline\hline
\multirow{40}{*}{\textbf{CD(S)R}}                 & \multirow{5}{*}{\textbf{BiTGCF}}     & HR@5             & 0.3112          & 0.1857          & 0.4487          & 0.2121          & 0.1997          & 0.9557          & 0.6039          & 0.6023          & 0.8543          & 0.3907           \\
                                              &                                      & HR@10            & 0.4287          & 0.2651          & 0.5640          & 0.3017          & 0.2807          & 0.9786          & 0.7449          & 0.7432          & 0.9168          & 0.5075           \\
                                              &                                      & NDCG@5           & 0.2196          & 0.1348          & 0.3410          & 0.1542          & 0.1451          & 0.8657          & 0.4464          & 0.4541          & 0.7537          & \underline{0.3025}   \\
                                              &                                      & NDCG@10          & 0.2575          & 0.1602          & 0.3784          & 0.1831          & 0.1712          & 0.8733          & 0.4922          & 0.4998          & 0.7743          & 0.3402           \\
                                              &                                      & MRR              & 0.2051          & 0.1284          & 0.3209          & 0.1471          & 0.1394          & 0.8733          & 0.4133          & 0.4239          & 0.7288          & 0.2890           \\ 
\cline{2-13}
                                              & \multirow{5}{*}{\textbf{DTCDR}}      & HR@5             & 0.2486          & 0.1881          & 0.4376          & 0.2005          & 0.2314          & 0.9435          & 0.6294          & 0.6076          & 0.7809          & 0.2556           \\
                                              &                                      & HR@10            & 0.3385          & 0.2875          & 0.5693          & 0.2972          & 0.3382          & 0.9745          & 0.7730          & 0.7623          & 0.9034          & 0.4341           \\
                                              &                                      & NDCG@5           & 0.1784          & 0.1224          & 0.3099          & 0.1406          & 0.1601          & 0.8414          & 0.4750          & 0.4680          & 0.5257          & 0.1591           \\
                                              &                                      & NDCG@10          & 0.2072          & 0.1544          & 0.3525          & 0.1716          & 0.1944          & 0.8516          & 0.5216          & 0.5181          & 0.5671          & 0.2163           \\
                                              &                                      & MRR              & 0.1670          & 0.1140          & 0.2854          & 0.1335          & 0.1507          & 0.8111          & 0.4432          & 0.5181          & 0.4596          & 0.1509           \\ 
\cline{2-13}
                                              & \multirow{5}{*}{\textbf{CMF}}        & HR@5             & 0.2020          & 0.1353          & 0.3764          & 0.1590          & 0.1500          & \textbf{0.9632} & 0.6198          & 0.6164          & 0.8519          & 0.3154           \\
                                              &                                      & HR@10            & 0.2918          & 0.1972          & 0.4990          & 0.2263          & 0.2089          & \textbf{0.9819} & 0.7675          & 0.7589          & 0.9151          & 0.4327           \\
                                              &                                      & NDCG@5           & 0.1442          & 0.0982          & 0.2796          & 0.1170          & 0.1106          & \textbf{0.8832} & 0.4743          & 0.4944          & 0.7650          & 0.2496           \\
                                              &                                      & NDCG@10          & 0.1731          & 0.1181          & 0.3193          & 0.1387          & 0.1295          & 0.8894          & 0.5222          & 0.5404          & 0.7859          & 0.2872           \\
                                              &                                      & MRR              & 0.1371          & 0.0942          & 0.2642          & 0.1122          & 0.1054          & 0.8585          & 0.4461          & 0.4731          & 0.7451          & 0.2435           \\ 
\cline{2-13}
                                              & \multirow{5}{*}{\textbf{CLFM}}       & HR@5             & 0.1994          & 0.1185          & 0.3712          & 0.1447          & 0.1319          & \underline{0.9607}  & 0.6579          & 0.5846          & 0.8241          & 0.2723           \\
                                              &                                     & HR@10            & 0.2830          & 0.1752          & 0.2677          & 0.2130          & 0.1878          & \underline{0.9801}  & 0.7950          & 0.7265          & 0.8885          & 0.3684           \\
                                              &                                      & NDCG@5           & 0.1430          & 0.0855          & 0.2677          & 0.1054          & 0.0968          & 0.8765          & 0.4982          & 0.4664          & 0.7320          & 0.2173           \\
                                              &                                      & NDCG@10          & 0.1698          & 0.1037          & 0.3106          & 0.1274          & 0.1147          & 0.8829          & 0.5428          & 0.5124          & 0.7531          & 0.2481           \\
                                              &                                      & MRR              & 0.1371          & 0.0942          & 0.2642          & 0.1122          & 0.1054          & 0.8585          & 0.4461          & 0.4731          & 0.7451          & 0.2435           \\ 
\cline{2-13}
                                              & \multirow{5}{*}{\textbf{CAT-ART}}    & HR@5             & \underline{0.3376}  & 0.2817          & 0.5128          & 0.2209          & 0.2994          & 0.9190          & 0.6376          & 0.6269          & 0.8568          & 0.3881           \\
                                              &                                      & HR@10            & \underline{0.4594}  & 0.2817          & 0.5919          & 0.3185          & 0.3904          & 0.9716          & 0.8083          & 0.7769          & 0.9272          & 0.5116           \\
                                              &                                      & NDCG@5           & 0.2265          & 0.1883          & 0.4047          & 0.1901          & 0.1937          & 0.8212          & 0.5174          & 0.5082          & 0.7629          & 0.2966           \\
                                              &                                      & NDCG@10          & \underline{0.2783}  & 0.2268          & 0.4237          & 0.2297          & 0.2371          & 0.8354          & 0.5301          & 0.5499          & 0.7727          & 0.3362           \\
                                              &                                      & MRR              & \textbf{0.2604} & 0.2098          & 0.3966          & 0.1958          & \underline{0.2205}  & 0.7996          & 0.4787          & 0.4545          & 0.7227          & 0.2918           \\ 
\cline{2-13}
                                              & \multirow{5}{*}{\textbf{C$^{2}$DSR}} & HR@5             & 0.2661          & 0.2022          & 0.4433          & 0.2256          & 0.2230          & 0.8582          & 0.6192          & 0.8281          & 0.8722          & 0.2510           \\
                                              &                                      & HR@10            & 0.3758          & 0.2877          & 0.5475          & 0.3186          & 0.3156          & 0.9052          & 0.7593          & 0.9142          & 0.9130          & 0.4943           \\
                                              &                                      & NDCG@5           & 0.1961          & 0.1497          & 0.3542          & 0.1618          & 0.1618          & 0.7598          & 0.4326          & 0.6411          & 0.6782          & 0.1347           \\
                                              &                                      & NDCG@10          & 0.2314          & 0.1796          & 0.3868          & 0.1917          & 0.1915          & 0.7751          & 0.4783          & 0.6696          & 0.6918          & 0.2133           \\
                                              &                                      & MRR              & 0.2100          & 0.1658          & 0.3458          & 0.1748          & 0.1751          & 0.7370          & 0.3994          & 0.5963          & 0.6201          & 0.1498           \\ 
\cline{2-13}
                                              & \multirow{5}{*}{\textbf{DeepAFP}}    & HR@5             & 0.1362          & 0.0956          & 0.2391          & 0.1239          & 0.1188          & 0.9592          & 0.5919          & 0.6109          & 0.8346          & 0.2770           \\
                                              &                                      & HR@10            & 0.2068          & 0.1500          & 0.3124          & 0.1790          & 0.1713          & 0.9767          & 0.7463          & 0.7539          & 0.8954          & 0.3719           \\
                                              &                                      & NDCG@5           & 0.0946          & 0.0664          & 0.1790          & 0.0925          & 0.0871          & \underline{0.8829}  & 0.4529          & 0.4803          & 0.7508          & 0.2245           \\
                                              &                                      & NDCG@10          & 0.1173          & 0.0838          & 0.2026          & 0.1102          & 0.1040          & 0.8878          & 0.5029          & 0.5199          & 0.7708          & 0.2540           \\
                                              &                                      & MRR              & 0.0903          & 0.0639          & 0.1688          & 0.0894          & 0.0837          & 0.8672          & 0.4278          & 0.4703          & 0.7315          & 0.2199           \\ 
\hhline{~============} 
                                              & \multirow{5}{*}{\textbf{CGRec}}                & HR@5             & \textbf{0.3476} & \textbf{0.3167} & \underline{0.5546}  & \textbf{0.3137} & \textbf{0.3162} & 0.9537          & \textbf{0.7839} & \textbf{0.9151} & \textbf{0.9262} & \textbf{0.4911}  \\
                                              &                                      & HR@10            & \textbf{0.4607} & \underline{0.4303}  & \underline{0.6423}  & \textbf{0.4228} & \textbf{0.4217} & 0.9784          & \textbf{0.8905} & 0.9712          & \underline{0.9736}  & \textbf{0.6566}  \\
                                              &                                      & NDCG@5           & \textbf{0.2564} & \textbf{0.2310} & \underline{0.4451}  & \textbf{0.2321} & \textbf{0.2330} & 0.8819          & \textbf{0.6279} & \underline{0.7850}  & 0.7882          & \textbf{0.3722}  \\
                                              &                                      & NDCG@10          & \textbf{0.2930} & \textbf{0.2677} & \underline{0.4732}  & \textbf{0.2672} & \textbf{0.2671} & \textbf{0.8900} & \textbf{0.6627} & \textbf{0.8034} & 0.8039          & \textbf{0.4256}  \\
                                              &                                      & MRR              & \underline{0.2595}  & \textbf{0.2373} & \underline{0.4330}  & \textbf{0.2392} & \textbf{0.2380} & \textbf{0.8622} & \textbf{0.5963} & \underline{0.7508}  & \underline{0.7500}  & \textbf{0.3732}  \\
\hline\hline
\end{tabular}
  \begin{tablenotes}
    \item[*] *In the Methods column, TR, SR, and CD(S)R indicate the traditional recommendation, sequential recommendation, and Cross-Domain (Sequential) Recommendation. Note that BiTGCF, DTCDR, CMF, CLFM, CAT-ART, DeepAPF belong to the CDR method, and  C$^{2}$DSR and our proposed model CGRec is the CDSR method.
    \end{tablenotes}
\end{table*}

\vspace{-0.19cm}
\subsection{Performance Comparisons (RQ1)}
\begin{figure}[ht]
    \begin{center}
    \includegraphics[width=0.98 \linewidth]{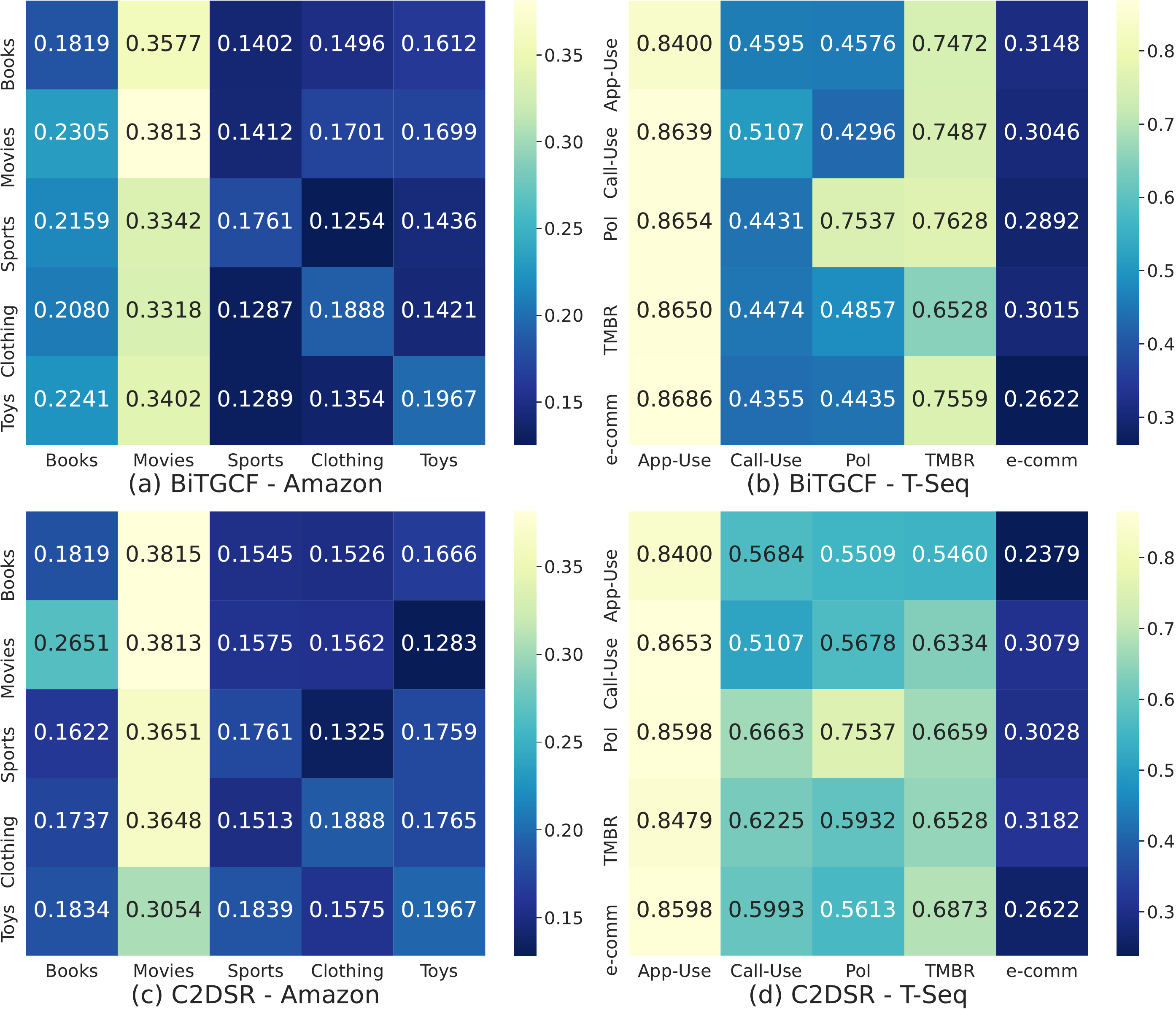}
    \end{center}
    \vspace{-0.3cm}
    \caption{NDCG@5 of Domain-Domain pairs in BiTGCF \cite{liu2020cross} and C$^{2}$DSR \cite{cao2022contrastive}.
    }
    \label{fig:result_2}
    \end{figure}

Table \ref{tab:result_1} shows the performance of our model compared to most of the CD(S)R and SR baseline methods within the cross-domain scenario, which comprises five domains. 
Since all CD(S)R models only modeled pairwise domain-domain relationships, two domains out of five domains were selected to train these models. 
Each domain was paired with four other domains, and the average of the performance of these four cases per domain is reported in the Table \ref{tab:result_1}.
Due to space limitations, a more detailed domain pair-wise performance of the CD(S)R models cannot be reported and only the results for BiTGCF and C$^{2}$DSR are shown in Figure \ref{fig:result_2}.

    \begin{figure*}[h]
    \begin{center}
    \includegraphics[width=0.92 \linewidth]{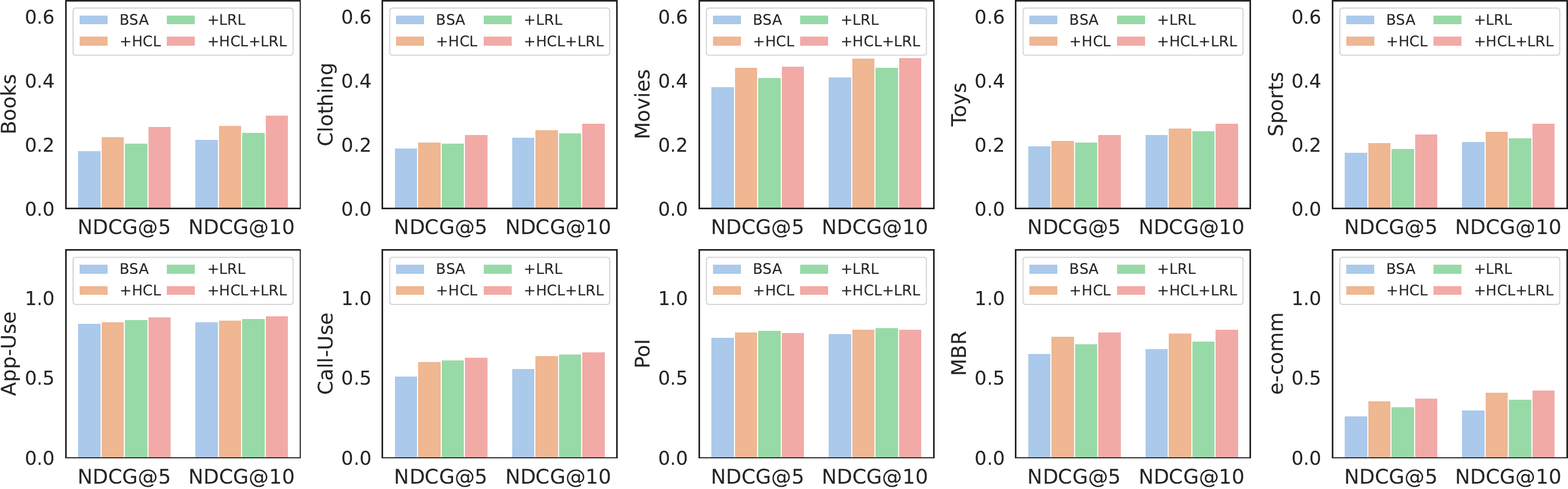}
    \end{center}
    \vspace{-0.3cm}
    \caption{Comparison of performance for different combinations of components.
    }
    \label{fig:ablation}
    \end{figure*}

Firstly, the CDR models performed poorly overall. 
There are two main reasons: (1) These models did not reflect the latest interests of users, as they failed to account for the dynamic sequential nature of user behavior; (2) Due to the negative transfer issue, these models tended to perform significantly worse for certain domain pairs in the experiments. 
As shown in Figure \ref{fig:result_2}a, the pairs of \textit{Sports} vs \textit{Books}, \textit{Movies}, \textit{Clothing}, \textit{Toys} domains in BiTGCF have performance deviations from 0.1287 to 0.1761 based on NDCG@5. 
In other words, the \textit{Sports} domain is being distorted by the \textit{Clothing} domain, resulting in a degradation of performance.
On the other hand, CAT-ART, which is the only CDR framework to model relationships between four or more domains, showed relatively reasonable performance. 
The reason is that the model did not reflect the temporal dynamics of user interaction, but considered the negative transfer of multiple domains in their model. 
This result suggests that expanding to three or more domains enhances the ability to reflect user preferences.

Secondly, the CDSR model, C$^{2}$DSR, performed relatively well, but not as well as CAT-ART, because of its inability to resolve the negative transfer between domains. 
In fact, C$^{2}$DSR, like the BiTGCF model, showed significant variation in performance between domain pairs. 
In C$^{2}$DSR, pairs of \textit{Sports} vs \textit{Books}, \textit{Movies}, \textit{Clothing}, \textit{Toys} domains have performance deviations from 0.1513 to 0.1839 based on NDCG@5, as shown in Figure \ref{fig:result_2}c. 
Our model outperformed the CD(S)R baseline for most of the domains in both datasets (RQ1), because it minimized the negative transfer between dissimilar domains, modelling the relationships of all five domains together. 
Our model also outperformed CAT-ART, which attempted to reduce the negative transfer across three or more domains, because our model incorporated the temporal dynamics of users into the model. 
In the CDSR task, it is crucial to incorporate temporal dynamics into the model. 
This is due to the ever-evolving nature of users' preferences, which consequently amplifies the dynamics of negative transfers.

Lastly, the performance of SR models was also measured using domain-hybrid sequences.
Overall, the existing SR models outperformed the CDR models. 
This indicates that the ability to capture the sequential nature of interactions provides valuable insights for making accurate recommendations.
However, the performance of SR models was still lower than our model because of the negative transfer issue between different domains. 
In particular, in the case of CORE, it is clear the performance of a specific domain (e.g., \textit{Sports}) is significantly lower. 
However, models that used category information as an additional input, such as FDSA, S$^{3}$Rec, and CARCA, showed relatively high performance compared to other SR models.
This suggests that the more general user preferences, such as category information, are strongly recommended to be incorporated in the model to mitigate the negative transfer issue. 
In comparison to these SR models, our model exhibited high performance across most of domains in both datasets (RQ1). 
This superior performance is attributed to taking into account the negative transfer between multiple domains.

\vspace{-0.3cm}
\subsection{Discussion of the negative transfer (RQ2)}
All CD(S)R and SR models examined in our study were significantly affected by the negative transfer issue, resulting in the performance degradation in several domains. 
This phenomenon was particularly evident in the \textit{Clothing} and \textit{Sports} domains of the \textit{Amazon} dataset, and in the \textit{Call-Use} and \textit{e-comm} domains of the \textit{T-Seq} dataset.
Our model showed higher performance gains for these domains compared to other domains as shown in Table \ref{tab:result_2}. 
The model outperformed the averaged NDCG@5 of all baselines of the \textit{Clothing} and \textit{Sports} domains in the \textit{Amazon} dataset, with gains of $\times 1.49$  and $\times 1.58$ times respectively.
Similarly, our model excelled in the \textit{Call-Use} and \textit{e-comm} domains in the \textit{T-Seq} dataset, with NDCG@5 gains of $\times 1.28$  and $\times 1.77$ times compared to the average of all baselines.
In essence, our model significantly improved the performance of domains that were negatively impacted by other domains. 
These results demonstrated that our model effectively solves the negative transfer issue, which contributes to its superior performance (RQ2).

\begin{table}[] \small
    \caption{Ratio of our model performance to the average performance of all baselines. The top two domains per dataset are highlighted in bold and underlined.}\label{tab:result_2}
    \vspace{-0.35cm}
\begin{tabular}{l|ccccc}
\hline\hline
\multicolumn{1}{c|}{\textbf{Metric}} & \textbf{Books} & \textbf{Clothing} & \textbf{Movies} & \textbf{Toys} & \textbf{Sports} \\ \hline
HR@5                                 & $\times$1.31           & \underline{$\times$1.44}              & $\times$1.32            & $\times$1.36          & \textbf{$\times$1.51}            \\
HR@10                                & $\times$1.25           & \underline{$\times$1.39}              & $\times$1.25            & $\times$1.30          & \textbf{$\times$1.43}            \\
NDCG@5                               & $\times$1.36           & \underline{$\times$1.49}              & $\times$1.35            & $\times$1.39          & \textbf{$\times$1.58}            \\
NDCG@10                              & $\times$1.32           & \underline{$\times$1.44}              & $\times$1.31            & $\times$1.35          & \textbf{$\times$1.52}            \\
MRR                                  & $\times$1.40           & \underline{$\times$1.53}              & $\times$1.37            & $\times$1.45          & \textbf{$\times$1.62}            \\ \hline\hline
\multicolumn{1}{c|}{\textbf{Metric}} & \textbf{App-Use}        & \textbf{Call-Use} & \textbf{PoI}    & \textbf{MBR}  & \textbf{e-comm} \\ \hline
HR@5                                 & $\times$1.02           & \underline{$\times$1.20}              & $\times$1.18            & $\times$1.08          & \textbf{$\times$1.61}            \\
HR@10                                & $\times$1.01           & \underline{$\times$1.12}              & $\times$1.11            & $\times$1.05          & \textbf{$\times$1.44}            \\
NDCG@5                               & $\times$1.05           & \underline{$\times$1.28}              & $\times$1.27            & $\times$1.13          & \textbf{$\times$1.72}            \\
NDCG@10                              & $\times$1.04           & \underline{$\times$1.24}              & $\times$1.23            & $\times$1.12          & \textbf{$\times$1.60}            \\
MRR                                  & $\times$1.05           & \underline{$\times$1.31}              & $\times$1.27            & $\times$1.14          & \textbf{$\times$1.72}            \\ \hline\hline
\end{tabular}
\end{table}

\vspace{-0.29cm}
\subsection{Discussion of Model Variants (RQ3)}
We carried out the following ablation studies to highlight the efficacy of each proposed module:

\noindent\textbf{(1) Basic Self-Attention model (BSA)}: The self-attention based model for the sequential recommendation, identical to SASRec except for the temporal encoding.

\noindent\textbf{(2) +Hierarchical Contrastive Learning (HCL)}: We changed the objective of the BSA model into the hierarchical contrastive loss.

\noindent\textbf{(3) +Loss Re-balancing Layer (LRL)}: We added loss re-balancing layer on the BSA.

Each of our modules improved the performance of the BSA model in all domains, and the model with both modules (+HCL+LRL) performed best in most domains, as shown in Figure \ref{fig:ablation}  (RQ3).
Specifically, the HCL module addressed the the negative transfer issue. 
The performance disparity between BSA and +HCL was most apparent in domains marked by the low performance domains, such as \textit{Sports} and \textit{e-comm} domain. 
In addition, the LRL also helped mitigate the negative transfer issue, as it outperformed the BSA model across all domains.
This phenomenon is particularly pronounced in the \textit{T-Seq} dataset, where there is a significant disparity in performance between domains. 

\vspace{-0.15cm}
\section{Conclusion}
This study focuses on the CDSR, a method using data from multiple domains for precise recommendations while recognizing users' dynamic interests. 
Negative transfer, a key challenge, is addressed by our proposed model, which quantifies and adaptively weights negative transfer in each domain. 
Furthermore, the model employed a hierarchical contrastive learning approach, capturing generalized user preferences. 
As a result, our model outperformed other CD(S)R and SR models on real-world datasets.

\vspace{-0.15cm}
\begin{acks}
This work was supported by Institute for Information \& communications Technology Planning \& Evaluation(IITP) grant funded by the Korea government(MSIT) (No. 2020-0-00368, A Neural-Symbolic Model for Knowledge Acquisition and Inference Techniques), Institute of Information \& communications Technology Planning \& Evaluation (IITP) grant funded by the Korea government (MSIT) (No.2019-0-00075, Artificial Intelligence Graduate School Program (KAIST)), and the National Research Foundation of Korea (NRF) grant funded by the Korea government (MSIT) (No. NRF-2022R1A2B \\ 5B02001913).
The authors would like to thank the AI Service Business Division of SK Telecom for providing GPU cluster support.
\end{acks}

\clearpage
\bibliographystyle{ACM-Reference-Format}
\balance
\bibliography{my_ref}


\begin{thebibliography}{27}


\ifx \showCODEN    \undefined \def \showCODEN     #1{\unskip}     \fi
\ifx \showDOI      \undefined \def \showDOI       #1{#1}\fi
\ifx \showISBNx    \undefined \def \showISBNx     #1{\unskip}     \fi
\ifx \showISBNxiii \undefined \def \showISBNxiii  #1{\unskip}     \fi
\ifx \showISSN     \undefined \def \showISSN      #1{\unskip}     \fi
\ifx \showLCCN     \undefined \def \showLCCN      #1{\unskip}     \fi
\ifx \shownote     \undefined \def \shownote      #1{#1}          \fi
\ifx \showarticletitle \undefined \def \showarticletitle #1{#1}   \fi
\ifx \showURL      \undefined \def \showURL       {\relax}        \fi
\providecommand\bibfield[2]{#2}
\providecommand\bibinfo[2]{#2}
\providecommand\natexlab[1]{#1}
\providecommand\showeprint[2][]{arXiv:#2}

\bibitem[Cao et~al\mbox{.}(2022)]%
        {cao2022contrastive}
\bibfield{author}{\bibinfo{person}{Jiangxia Cao}, \bibinfo{person}{Xin Cong}, \bibinfo{person}{Jiawei Sheng}, \bibinfo{person}{Tingwen Liu}, {and} \bibinfo{person}{Bin Wang}.} \bibinfo{year}{2022}\natexlab{}.
\newblock \showarticletitle{Contrastive Cross-Domain Sequential Recommendation}. In \bibinfo{booktitle}{\emph{Proceedings of the 31st ACM International Conference on Information \& Knowledge Management}}. \bibinfo{pages}{138--147}.
\newblock


\bibitem[Gao et~al\mbox{.}(2013)]%
        {gao2013cross}
\bibfield{author}{\bibinfo{person}{Sheng Gao}, \bibinfo{person}{Hao Luo}, \bibinfo{person}{Da Chen}, \bibinfo{person}{Shantao Li}, \bibinfo{person}{Patrick Gallinari}, {and} \bibinfo{person}{Jun Guo}.} \bibinfo{year}{2013}\natexlab{}.
\newblock \showarticletitle{Cross-domain recommendation via cluster-level latent factor model}. In \bibinfo{booktitle}{\emph{Machine Learning and Knowledge Discovery in Databases: European Conference, ECML PKDD 2013, Prague, Czech Republic, September 23-27, 2013, Proceedings, Part II 13}}. Springer, \bibinfo{pages}{161--176}.
\newblock


\bibitem[Hendrycks and Gimpel(2016)]%
        {hendrycks2016gaussian}
\bibfield{author}{\bibinfo{person}{Dan Hendrycks} {and} \bibinfo{person}{Kevin Gimpel}.} \bibinfo{year}{2016}\natexlab{}.
\newblock \showarticletitle{Gaussian error linear units (gelus)}.
\newblock \bibinfo{journal}{\emph{arXiv preprint arXiv:1606.08415}} (\bibinfo{year}{2016}).
\newblock


\bibitem[Hidasi et~al\mbox{.}(2015)]%
        {hidasi2015session}
\bibfield{author}{\bibinfo{person}{Bal{\'a}zs Hidasi}, \bibinfo{person}{Alexandros Karatzoglou}, \bibinfo{person}{Linas Baltrunas}, {and} \bibinfo{person}{Domonkos Tikk}.} \bibinfo{year}{2015}\natexlab{}.
\newblock \showarticletitle{Session-based recommendations with recurrent neural networks}.
\newblock \bibinfo{journal}{\emph{arXiv preprint arXiv:1511.06939}} (\bibinfo{year}{2015}).
\newblock


\bibitem[Hou et~al\mbox{.}(2022)]%
        {hou2022core}
\bibfield{author}{\bibinfo{person}{Yupeng Hou}, \bibinfo{person}{Binbin Hu}, \bibinfo{person}{Zhiqiang Zhang}, {and} \bibinfo{person}{Wayne~Xin Zhao}.} \bibinfo{year}{2022}\natexlab{}.
\newblock \showarticletitle{Core: simple and effective session-based recommendation within consistent representation space}. In \bibinfo{booktitle}{\emph{Proceedings of the 45th international ACM SIGIR conference on research and development in information retrieval}}. \bibinfo{pages}{1796--1801}.
\newblock


\bibitem[Kang and McAuley(2018)]%
        {kang2018self}
\bibfield{author}{\bibinfo{person}{Wang-Cheng Kang} {and} \bibinfo{person}{Julian McAuley}.} \bibinfo{year}{2018}\natexlab{}.
\newblock \showarticletitle{Self-attentive sequential recommendation}. In \bibinfo{booktitle}{\emph{2018 IEEE international conference on data mining (ICDM)}}. IEEE, \bibinfo{pages}{197--206}.
\newblock


\bibitem[Li et~al\mbox{.}(2023)]%
        {li2023one}
\bibfield{author}{\bibinfo{person}{Chenglin Li}, \bibinfo{person}{Yuanzhen Xie}, \bibinfo{person}{Chenyun Yu}, \bibinfo{person}{Bo Hu}, \bibinfo{person}{Zang Li}, \bibinfo{person}{Guoqiang Shu}, \bibinfo{person}{Xiaohu Qie}, {and} \bibinfo{person}{Di Niu}.} \bibinfo{year}{2023}\natexlab{}.
\newblock \showarticletitle{One for All, All for One: Learning and Transferring User Embeddings for Cross-Domain Recommendation}. In \bibinfo{booktitle}{\emph{Proceedings of the Sixteenth ACM International Conference on Web Search and Data Mining}}. \bibinfo{pages}{366--374}.
\newblock


\bibitem[Liu et~al\mbox{.}(2020)]%
        {liu2020cross}
\bibfield{author}{\bibinfo{person}{Meng Liu}, \bibinfo{person}{Jianjun Li}, \bibinfo{person}{Guohui Li}, {and} \bibinfo{person}{Peng Pan}.} \bibinfo{year}{2020}\natexlab{}.
\newblock \showarticletitle{Cross domain recommendation via bi-directional transfer graph collaborative filtering networks}. In \bibinfo{booktitle}{\emph{Proceedings of the 29th ACM international conference on information \& knowledge management}}. \bibinfo{pages}{885--894}.
\newblock


\bibitem[Ma et~al\mbox{.}(2019)]%
        {ma2019pi}
\bibfield{author}{\bibinfo{person}{Muyang Ma}, \bibinfo{person}{Pengjie Ren}, \bibinfo{person}{Yujie Lin}, \bibinfo{person}{Zhumin Chen}, \bibinfo{person}{Jun Ma}, {and} \bibinfo{person}{Maarten~de Rijke}.} \bibinfo{year}{2019}\natexlab{}.
\newblock \showarticletitle{$\pi$-net: A parallel information-sharing network for shared-account cross-domain sequential recommendations}. In \bibinfo{booktitle}{\emph{Proceedings of the 42nd international ACM SIGIR conference on research and development in information retrieval}}. \bibinfo{pages}{685--694}.
\newblock


\bibitem[McAuley et~al\mbox{.}(2015)]%
        {mcauley2015image}
\bibfield{author}{\bibinfo{person}{Julian McAuley}, \bibinfo{person}{Christopher Targett}, \bibinfo{person}{Qinfeng Shi}, {and} \bibinfo{person}{Anton Van Den~Hengel}.} \bibinfo{year}{2015}\natexlab{}.
\newblock \showarticletitle{Image-based recommendations on styles and substitutes}. In \bibinfo{booktitle}{\emph{Proceedings of the 38th international ACM SIGIR conference on research and development in information retrieval}}. \bibinfo{pages}{43--52}.
\newblock


\bibitem[Myerson(1991)]%
        {myerson1991game}
\bibfield{author}{\bibinfo{person}{Roger~B Myerson}.} \bibinfo{year}{1991}\natexlab{}.
\newblock \bibinfo{booktitle}{\emph{Game theory: analysis of conflict}}.
\newblock \bibinfo{publisher}{Harvard university press}.
\newblock


\bibitem[Rashed et~al\mbox{.}(2022)]%
        {rashed2022context}
\bibfield{author}{\bibinfo{person}{Ahmed Rashed}, \bibinfo{person}{Shereen Elsayed}, {and} \bibinfo{person}{Lars Schmidt-Thieme}.} \bibinfo{year}{2022}\natexlab{}.
\newblock \showarticletitle{Context and attribute-aware sequential recommendation via cross-attention}. In \bibinfo{booktitle}{\emph{Proceedings of the 16th ACM Conference on Recommender Systems}}. \bibinfo{pages}{71--80}.
\newblock


\bibitem[Rendle et~al\mbox{.}(2012)]%
        {rendle2012bpr}
\bibfield{author}{\bibinfo{person}{Steffen Rendle}, \bibinfo{person}{Christoph Freudenthaler}, \bibinfo{person}{Zeno Gantner}, {and} \bibinfo{person}{Lars Schmidt-Thieme}.} \bibinfo{year}{2012}\natexlab{}.
\newblock \showarticletitle{BPR: Bayesian personalized ranking from implicit feedback}.
\newblock \bibinfo{journal}{\emph{arXiv preprint arXiv:1205.2618}} (\bibinfo{year}{2012}).
\newblock


\bibitem[Rendle et~al\mbox{.}(2010)]%
        {rendle2010factorizing}
\bibfield{author}{\bibinfo{person}{Steffen Rendle}, \bibinfo{person}{Christoph Freudenthaler}, {and} \bibinfo{person}{Lars Schmidt-Thieme}.} \bibinfo{year}{2010}\natexlab{}.
\newblock \showarticletitle{Factorizing personalized markov chains for next-basket recommendation}. In \bibinfo{booktitle}{\emph{Proceedings of the 19th international conference on World wide web}}. \bibinfo{pages}{811--820}.
\newblock


\bibitem[Singh and Gordon(2008)]%
        {singh2008relational}
\bibfield{author}{\bibinfo{person}{Ajit~P Singh} {and} \bibinfo{person}{Geoffrey~J Gordon}.} \bibinfo{year}{2008}\natexlab{}.
\newblock \showarticletitle{Relational learning via collective matrix factorization}. In \bibinfo{booktitle}{\emph{Proceedings of the 14th ACM SIGKDD international conference on Knowledge discovery and data mining}}. \bibinfo{pages}{650--658}.
\newblock


\bibitem[Straffin(1993)]%
        {straffin1993game}
\bibfield{author}{\bibinfo{person}{Philip~D Straffin}.} \bibinfo{year}{1993}\natexlab{}.
\newblock \bibinfo{booktitle}{\emph{Game theory and strategy}}. Vol.~\bibinfo{volume}{36}.
\newblock \bibinfo{publisher}{MAA}.
\newblock


\bibitem[Sun et~al\mbox{.}(2019)]%
        {sun2019bert4rec}
\bibfield{author}{\bibinfo{person}{Fei Sun}, \bibinfo{person}{Jun Liu}, \bibinfo{person}{Jian Wu}, \bibinfo{person}{Changhua Pei}, \bibinfo{person}{Xiao Lin}, \bibinfo{person}{Wenwu Ou}, {and} \bibinfo{person}{Peng Jiang}.} \bibinfo{year}{2019}\natexlab{}.
\newblock \showarticletitle{BERT4Rec: Sequential recommendation with bidirectional encoder representations from transformer}. In \bibinfo{booktitle}{\emph{Proceedings of the 28th ACM international conference on information and knowledge management}}. \bibinfo{pages}{1441--1450}.
\newblock


\bibitem[Szegedy et~al\mbox{.}(2016)]%
        {szegedy2016rethinking}
\bibfield{author}{\bibinfo{person}{Christian Szegedy}, \bibinfo{person}{Vincent Vanhoucke}, \bibinfo{person}{Sergey Ioffe}, \bibinfo{person}{Jon Shlens}, {and} \bibinfo{person}{Zbigniew Wojna}.} \bibinfo{year}{2016}\natexlab{}.
\newblock \showarticletitle{Rethinking the inception architecture for computer vision}. In \bibinfo{booktitle}{\emph{Proceedings of the IEEE conference on computer vision and pattern recognition}}. \bibinfo{pages}{2818--2826}.
\newblock


\bibitem[Vaswani et~al\mbox{.}(2017)]%
        {vaswani2017attention}
\bibfield{author}{\bibinfo{person}{Ashish Vaswani}, \bibinfo{person}{Noam Shazeer}, \bibinfo{person}{Niki Parmar}, \bibinfo{person}{Jakob Uszkoreit}, \bibinfo{person}{Llion Jones}, \bibinfo{person}{Aidan~N Gomez}, \bibinfo{person}{{\L}ukasz Kaiser}, {and} \bibinfo{person}{Illia Polosukhin}.} \bibinfo{year}{2017}\natexlab{}.
\newblock \showarticletitle{Attention is all you need}. In \bibinfo{booktitle}{\emph{Advances in neural information processing systems}}. \bibinfo{pages}{5998--6008}.
\newblock


\bibitem[Yan et~al\mbox{.}(2019)]%
        {yan2019deepapf}
\bibfield{author}{\bibinfo{person}{Huan Yan}, \bibinfo{person}{Xiangning Chen}, \bibinfo{person}{Chen Gao}, \bibinfo{person}{Yong Li}, {and} \bibinfo{person}{Depeng Jin}.} \bibinfo{year}{2019}\natexlab{}.
\newblock \showarticletitle{Deepapf: Deep attentive probabilistic factorization for multi-site video recommendation}.
\newblock \bibinfo{journal}{\emph{TC}} \bibinfo{volume}{2}, \bibinfo{number}{130} (\bibinfo{year}{2019}), \bibinfo{pages}{17--883}.
\newblock


\bibitem[Zhang et~al\mbox{.}(2019)]%
        {zhang2019feature}
\bibfield{author}{\bibinfo{person}{Tingting Zhang}, \bibinfo{person}{Pengpeng Zhao}, \bibinfo{person}{Yanchi Liu}, \bibinfo{person}{Victor~S Sheng}, \bibinfo{person}{Jiajie Xu}, \bibinfo{person}{Deqing Wang}, \bibinfo{person}{Guanfeng Liu}, \bibinfo{person}{Xiaofang Zhou}, {et~al\mbox{.}}} \bibinfo{year}{2019}\natexlab{}.
\newblock \showarticletitle{Feature-level Deeper Self-Attention Network for Sequential Recommendation.}. In \bibinfo{booktitle}{\emph{IJCAI}}. \bibinfo{pages}{4320--4326}.
\newblock


\bibitem[Zhang et~al\mbox{.}(2020)]%
        {zhang2020overcoming}
\bibfield{author}{\bibinfo{person}{Wen Zhang}, \bibinfo{person}{Lingfei Deng}, {and} \bibinfo{person}{Dongrui Wu}.} \bibinfo{year}{2020}\natexlab{}.
\newblock \showarticletitle{Overcoming negative transfer: A survey}.
\newblock \bibinfo{journal}{\emph{arXiv preprint arXiv:2009.00909}} (\bibinfo{year}{2020}), \bibinfo{pages}{36}.
\newblock


\bibitem[Zhao et~al\mbox{.}(2022)]%
        {recbole[2.0]}
\bibfield{author}{\bibinfo{person}{Wayne~Xin Zhao}, \bibinfo{person}{Yupeng Hou}, \bibinfo{person}{Xingyu Pan}, \bibinfo{person}{Chen Yang}, \bibinfo{person}{Zeyu Zhang}, \bibinfo{person}{Zihan Lin}, \bibinfo{person}{Jingsen Zhang}, \bibinfo{person}{Shuqing Bian}, \bibinfo{person}{Jiakai Tang}, \bibinfo{person}{Wenqi Sun}, {et~al\mbox{.}}} \bibinfo{year}{2022}\natexlab{}.
\newblock \showarticletitle{RecBole 2.0: Towards a More Up-to-Date Recommendation Library}. In \bibinfo{booktitle}{\emph{Proceedings of the 31st ACM International Conference on Information \& Knowledge Management}}. \bibinfo{pages}{4722--4726}.
\newblock


\bibitem[Zhao et~al\mbox{.}(2021)]%
        {zhao2021recbole}
\bibfield{author}{\bibinfo{person}{Wayne~Xin Zhao}, \bibinfo{person}{Shanlei Mu}, \bibinfo{person}{Yupeng Hou}, \bibinfo{person}{Zihan Lin}, \bibinfo{person}{Kaiyuan Li}, \bibinfo{person}{Yushuo Chen}, \bibinfo{person}{Yujie Lu}, \bibinfo{person}{Hui Wang}, \bibinfo{person}{Changxin Tian}, \bibinfo{person}{Xingyu Pan}, \bibinfo{person}{Yingqian Min}, \bibinfo{person}{Zhichao Feng}, \bibinfo{person}{Xinyan Fan}, \bibinfo{person}{Xu Chen}, \bibinfo{person}{Pengfei Wang}, \bibinfo{person}{Wendi Ji}, \bibinfo{person}{Yaliang Li}, \bibinfo{person}{Xiaoling Wang}, {and} \bibinfo{person}{Ji-Rong Wen}.} \bibinfo{year}{2021}\natexlab{}.
\newblock \showarticletitle{Recbole: Towards a unified, comprehensive and efficient framework for recommendation algorithms}. In \bibinfo{booktitle}{\emph{{CIKM}}}.
\newblock


\bibitem[Zhou et~al\mbox{.}(2020)]%
        {zhou2020s3}
\bibfield{author}{\bibinfo{person}{Kun Zhou}, \bibinfo{person}{Hui Wang}, \bibinfo{person}{Wayne~Xin Zhao}, \bibinfo{person}{Yutao Zhu}, \bibinfo{person}{Sirui Wang}, \bibinfo{person}{Fuzheng Zhang}, \bibinfo{person}{Zhongyuan Wang}, {and} \bibinfo{person}{Ji-Rong Wen}.} \bibinfo{year}{2020}\natexlab{}.
\newblock \showarticletitle{S3-rec: Self-supervised learning for sequential recommendation with mutual information maximization}. In \bibinfo{booktitle}{\emph{Proceedings of the 29th ACM international conference on information \& knowledge management}}. \bibinfo{pages}{1893--1902}.
\newblock


\bibitem[Zhu et~al\mbox{.}(2019)]%
        {zhu2019dtcdr}
\bibfield{author}{\bibinfo{person}{Feng Zhu}, \bibinfo{person}{Chaochao Chen}, \bibinfo{person}{Yan Wang}, \bibinfo{person}{Guanfeng Liu}, {and} \bibinfo{person}{Xiaolin Zheng}.} \bibinfo{year}{2019}\natexlab{}.
\newblock \showarticletitle{Dtcdr: A framework for dual-target cross-domain recommendation}. In \bibinfo{booktitle}{\emph{Proceedings of the 28th ACM International Conference on Information and Knowledge Management}}. \bibinfo{pages}{1533--1542}.
\newblock


\bibitem[Zhu et~al\mbox{.}(2021)]%
        {zhu2021cross}
\bibfield{author}{\bibinfo{person}{Feng Zhu}, \bibinfo{person}{Yan Wang}, \bibinfo{person}{Chaochao Chen}, \bibinfo{person}{Jun Zhou}, \bibinfo{person}{Longfei Li}, {and} \bibinfo{person}{Guanfeng Liu}.} \bibinfo{year}{2021}\natexlab{}.
\newblock \showarticletitle{Cross-domain recommendation: challenges, progress, and prospects}.
\newblock \bibinfo{journal}{\emph{arXiv preprint arXiv:2103.01696}} (\bibinfo{year}{2021}).
\newblock


\end{thebibliography}
\end{document}